\documentclass[journal]{IEEEtran}

\usepackage[linesnumbered,ruled]{algorithm2e}
\usepackage{amssymb}
\usepackage{amsmath}
\usepackage{caption}
\usepackage{xcolor}
\usepackage[pdftex]{graphicx}
\usepackage{graphics}
\usepackage[caption=false,font=footnotesize]{subfig}
\usepackage{cite}
\ifCLASSINFOpdf
   \usepackage[pdftex]{graphicx}
\else  
   \usepackage[dvips]{graphicx}  
\fi

\usepackage[linesnumbered,ruled]{algorithm2e}
\usepackage{amssymb}
\usepackage{xcolor}

\begin{document}
\title{Lateralization in Agents' Decision Making: Evidence of Benefits/Costs from Artificial Intelligence}

\author{Abubakar~Siddique,~\IEEEmembership{Member,~IEEE,}
        Will~N.~Browne,~\IEEEmembership{Member,~IEEE}
        and~Gina~M.~Grimshaw% <-this % stops a space
\thanks{Dr. Siddique is with the School of Engineering and Computer Science, Victoria University of Wellington, PO Box 600, Wellington 6140, New Zealand.}% <-this % stops a space
\thanks{Prof. Browne is with the Faculty of Engineering, School of Electrical Engineering \& Robotics, Queensland University of Technology, Brisbane 4000, Australia.}
\thanks{A/Prof. Grimshaw is with the Cognitive and Affective Neuroscience lab at the School of Psychology, Victoria University of Wellington, PO Box 600, Wellington 6140, New Zealand.}% <-this % stops a space
\thanks{Manuscript received -- --, ----; revised -- --, ----.}}

\markboth{}%
{Siddique \MakeLowercase{\textit{et al.}}: Lateralization in Agents' Decision Making: Evidence of Benefits/Costs from Artificial Intelligence}

\maketitle

%%Limits
%%%%%%%%%%%%%%%%%%%%%%%%%%%%%%%%%%%%%%%%%%%%%%%%%%%%%%%%%%%%%%%%%%%%%%%%%%%%%%%%%
%Pages              10
%Abstract           100-200 words
%Keywords           2-5
%Fontsize           10

\begin{abstract}
Lateralization is ubiquitous in vertebrate brains which, as well as its role in locomotion, is considered an important factor in biological intelligence. Lateralization has been associated with both poor and good performance. It has been hypothesized that lateralization has benefits that may counterbalance its costs. Given that lateralization is ubiquitous, it likely has advantages that can benefit artificial intelligence. In turn, lateralized artificial intelligent systems can be used as tools to advance the understanding of lateralization in biological intelligence. Recently lateralization has been incorporated into artificially intelligent systems to solve complex problems in computer vision and navigation domains. Here we describe and test two novel lateralized artificial intelligent systems that simultaneously represent and address given problems at constituent and holistic levels. The experimental results demonstrate that the lateralized systems outperformed state-of-the-art non-lateralized systems in resolving complex problems. The advantages arise from the abilities, (i) to represent an input signal at both the constituent level and holistic level simultaneously, such that the most appropriate viewpoint controls the system; (ii) to avoid extraneous computations by generating excite and inhibit signals. The computational costs associated with the lateralized AI systems are either less than the conventional AI systems or countered by providing better solutions.
\end{abstract}

% NOTe that keywords are not normally used for peerreview papers.

\IEEEpeerreviewmaketitle
%\vspace{-2.0em}
Biological intelligence has been used as a major source of inspiration for creating artificially intelligent (AI) systems \cite{mcculloch1943logical,hopfield1982neural,holland1998cognitive, siddique2022lateralized, siddique2021lateralizedthesis}. In turn, AI systems can be used as tools to advance understand of biological intelligence. These advances can be created in three main ways. First, AI systems can be employed as data mining tools to extract patterns from neuroscience data that are too complex for conventional statistical approaches to handle well. Second, AI can model the brain in order to replicate it. Third, AI systems can be used to probe fundamental aspects of cognitive architecture. It is this third approach we take here.

Given that lateralization is ubiquitous in brains, some evolutionary benefits can be assumed, at least in some domains. But that does not mean those benefits extend to all domains. The research community has been struggling to determine the trade-off between the benefits and costs of lateralization. It has been hypothesized that lateralization has benefits that may counterbalance its costs \cite{chivers2017competitive,bibost2014laterality,magat2009laterality}. Across domains, lateralization has been associated with both poor and good performance \cite{boles2008asymmetry,hirnstein2014brain}.

The relationship between lateralization and cognitive performance may depends on the task at hand. Some tasks show that better performance is associated with greater lateralization. For example, lateralized chicks have better performance to detect the model predator while searching for food as compared to the non-lateralized chicks \cite{rogers2004advantages,dharmaretnam2005hemispheric,rogers2000evolution}. However, lateralization is not universally advantageous, and may entail costs \cite{levy1972lateral}. It has been hypothesized that lateralization has benefits that may counterbalance its costs \cite{chivers2017competitive,bibost2014laterality,magat2009laterality}. The benefits of lateralization relative to its costs are still a debatable topic.

Handedness is a physical manifestation of lateralization so could provide insight into the relationship between lateralization and cognitive performance. It is the preferential use of one hand that is faster, better, more capable, and gives a more precise performance on manual tests \cite{holder2002does}. Although most (i.e., $90\%$) of humans are right-handed, there is individual variability in both the direction and degree of handedness. The relationship between handedness and cognitive abilities has been explored by several studies but is ambiguous yet \cite{hirnstein2019cognitive}. For example, cognitive task-based higher performance has been associated with the right-handers \cite{johnston2009nature,nicholls2010relationship}. However, the performance difference has also been associated with the strong and weak hand preference, rather than the left- or right-handers. That is the higher performance is associated with handedness strength rather than handedness direction \cite{corballis2008handedness,leask2006single}.

Similarly, the relationship between cognitive performance and behavioral lateralization (behaviourally assessed hemispheric asymmetry) is equivocal. The performance of verbal tasks, assessed with dichotic listening, is positively associated with the left-hemispheric language strength \cite{hirnstein2014brain,barth2012preschool}. In contrast, the performance of verbal tasks, assessed with visual half-field representation, is associated with the symmetric language representation \cite{hirnstein2014brain}. One speculation is that the functions that lateralized at the start and end of the ontogenetic development exhibit a positive laterality-performance correlation, whereas, the functions that lateralized at the intermediate stage exhibit a negative laterality-performance correlation \cite{boles2008asymmetry}.

Inter-hemispheric interaction is another factor that can be considered when investigating lateralization. Brain hemispheres primarily communicate with each other through the corpus callosum. The connections between them can be excitatory or inhibitory. The excitatory signals are important for the transmission of information and allow integration. Consequently, increased inter-hemispheric communication results in weaker independent lateralization. However, inhibitory signals are also important enabling one hemisphere to dominate processing depending on task goals. Consequently, increased inter-hemispheric communication can result in strong lateralization \cite{stark2008regional,grimshaw2015hemispheric}. 

From the above discussion, it is clear that although there is ample evidence that cognitive performance and lateralization are associated, there is inconsistency regarding the nature of this relationship, i.e. whether lateralization is positively or negatively correlated with cognitive performance, or has no effects. These inconsistencies may arise because of trade-offs between the costs and benefits of lateralization, and which dominate in a given task. Moreover, these inconsistent findings may reflect poor methodology (small sample sizes, noisy data). But they may also indicate that lateralization has both costs and benefits, and so the association between lateralization and performance will depend heavily on whether specific task parameters are biased towards benefits or costs.

Empirical evidence regarding the relationship between performance and lateralization could be obtained from AI systems. It is anticipated that the integration of lateralization into AI systems is beneficial/advantageous in certain situations. However, it is not clear what those situations are, what are the benefits/advantages, and what are the costs involved. This work will investigate lateralization in artificial agents' decision-making to obtain evidence of the trade-off between benefits and costs from artificial intelligence in order to inform cognitive neuroscience.

In this article, we describe two lateralized systems that we have developed, each designed to address a different type of problem. The first system is developed to evaluate whether lateralized approach functions well in complex computer vision problems (e.g., identifying animals in digital images) where conventional systems often fail. The second system is developed to test the effectiveness of the lateralized approach for solving complex navigation problems, where we know that vertebrate brains use multiple levels of representation simultaneously. The following sections provide a brief description of conventional (homogeneous) and lateralized AI systems, and present case studies for computer vision and navigation problems.

\begin{figure}
	\begin{center}
		\includegraphics [scale=0.26]{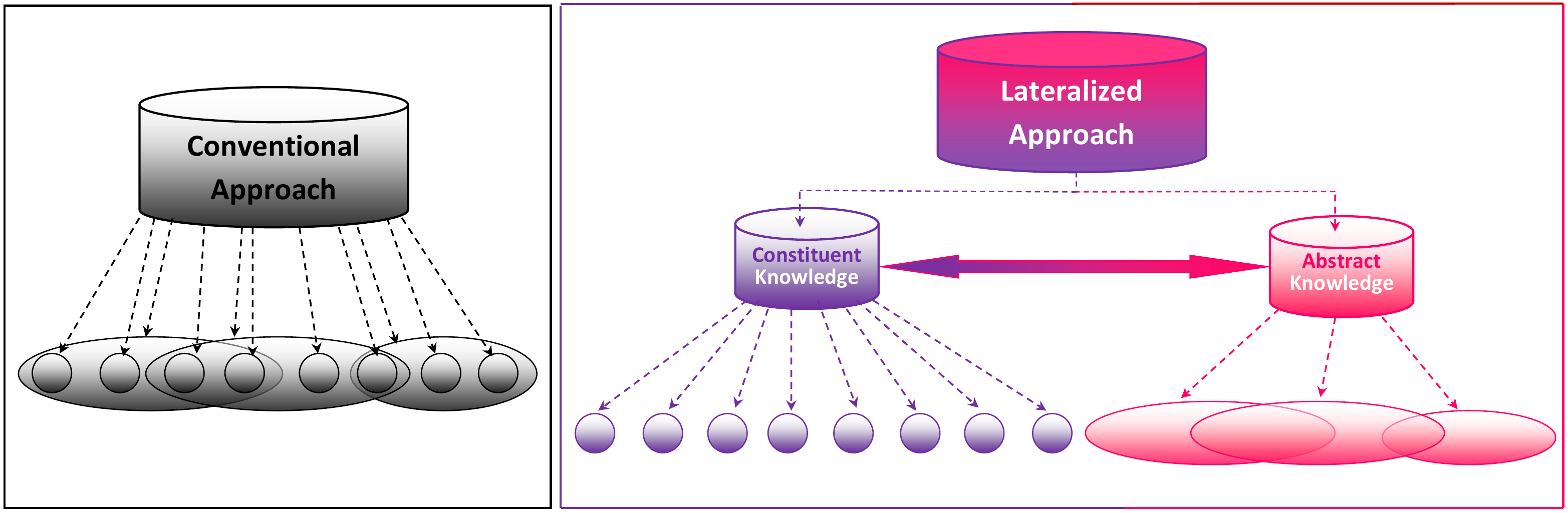}
		%\vspace{-0.70em}
		\caption{A schematic illustration of a conventional homogeneous approach and a lateralized approach. A conventional AI approach considers individual features and niches in a homogeneous manner, whereas, a lateralized approach splits a complex problem into constituent (left half) and abstract (right half) knowledge.}
		%\vspace{-2.4em}
		\label{ConvSysLatSys}
		%\vspace{-.60em}
	\end{center}
\end{figure}

\section{Homogeneous Vs Lateralized AI Systems}
The majority of AI systems are homogeneous. A homogeneous AI system is characterized by considering all features equally. Homogeneous systems cannot differentiate between constituent-level simple features and holistic-level complex features. Homogeneous systems work well when there are only simple features in the domain or the relationship between features and target (e.g. class or action) is linearly separable (i.e. a linear combination of features can be used to separate out specific targets). However, these systems struggle when there are complex patterns of features in the domain. Often these complex features are made-up of patterns of features, i.e. hierarchical patterns within patterns. The ability to consider the problem at multiple levels may effectively resolve such complex hierarchical features.

Artificially intelligent systems based on the principle of hemispheric lateralization are termed lateralized AI systems. Such a system can process an input signal at the constituent level and holistic level simultaneously. This enables lateralized AI systems to address both individual features and global patterns in parallel. Thus to create lateralized AI system, it is required to handle heterogeneous features, which represent sensed parts of the state of an environment. These heterogeneous features are used to construct a hierarchical representation of knowledge at different levels of abstraction, i.e. at a constituent level and holistic level, which can be split across hemispheres. Each knowledge component, a building block of knowledge (BBK)\footnote{A BBK is a unit of knowledge that is transferable and can be used or reused to solve a part of a problem.}, contains sufficient information to resolve a part of a problem \cite{konidaris2019necessity}.

To create a lateralized AI system, it needs to mimic the type of heterogeneity which can be seen in biological intelligence. It must be a modular system such that each module can solve a part of a problem. It needs the ability to learn knowledge from simple and small-scale problems and then (re)utilize it to learn complex and large-scale problems in related and similar domains. Two organizing principles of brains support this re-utilization of learned knowledge, i.e. lateralization (as discussed above) and modularity of function \cite{corballis2017evolution,krichmar2008neuromodulatory,alexander2015retrosplenial,nitz2012spaces}. Here modularity does not mean macro-level modules for whole cognitive functions, but micro-level modules that are building blocks of knowledge that can be (re)combined. Lateralization allows modular learning at different levels of abstraction \cite{corballis2017evolution,krichmar2008neuromodulatory}. This enables brains to process individual features and global patterns simultaneously. For instance, the left hemisphere generally processes elementary information (constituent level), whereas the right hemisphere generally processes the same information at a higher level of abstraction (holistic level) \cite{gaddes2013learning,luria2012higher,dharmaretnam2005hemispheric,rogers2004advantages}.

The incorporation of these principles in AI systems could empower them to overcome the limitations inherent in conventional (homogeneous) AI systems and provides robust and efficient solutions for complex hierarchical problems \cite{siddique2021lateralizedthesis}. The resultant lateralized AI systems should be able to generate knowledge representation at different levels of abstraction, i.e. constituent level and holistic level. One half of the system (say \textit{left} half) utilizes constituent BBKs to represent the basic elements of knowledge such as individual features and simple niches\footnote{The area of a sample space where neighboring instances share a common property is called a niche.}; whereas the other (\textit{right}) half of the system utilizes holistic BBKs to represent higher-order abstract features that are extracted across niches. These constituent and holistic BBKs provide a local viewpoint and a big picture of the overarching patterns in the data, respectively. Finally, the \textit{left} and \textit{right} half of the system communicate to efficiently solve the problem. A schematic illustration of conventional homogeneous and lateralized approaches is shown in Figure \ref{ConvSysLatSys}.

\section{Case Study 1: Lateralized AI System for Computer Vision Problems}
\label{CaseStudy1}
Computer vision problems are single-step visual classification tasks taken from the real-world domain. The ground-truth information is often available for these problems when humans label the data. Moreover, these problems include uncertainty, noise, and irrelevant and redundant data. These characteristics make computer vision problems ideal candidates with which to evaluate the robustness of the lateralized approach \cite{siddique2023lateralized}.

Instead of spike trains in the visual cortex, computer vision presents pixel intensities in red, green, and blue channels in an image of specific size. The following section presents an overview of the required background knowledge from the computer vision (Image features and adversarial attacks) and machine learning (deep learning and LCS) techniques that are relevant to this work. The computer vision techniques are used to generate an input signal for the novel AI system, whereas, machine learning techniques are used to create  the novel lateralized AI system.

\subsection{Background}
\label{BacGro1}

\subsubsection{Image Features}
\label{Feat}
An image feature is a BBK that provides important information about a specific aspect, pattern, structure, or characteristic of an image. For this work, two commonly used feature extraction techniques are applied to extract features from the images, i.e. scale-invariant feature transform and histogram of oriented gradients. The scale-invariant feature transform (SIFT) is a state-of-the-art technique that has been commonly used for object detection and image classification \cite{lowe2004distinctive}. The image measurements are performed by the SIFT descriptor in the form of receptive fields over which local scale-invariant reference frames are established by using local scale selection. The SIFT features are invariant to uniform scaling, illumination changes, and orientation \cite{lowe1999object,ke2004efficient}.

The histogram of oriented gradients (HOG) is another well-recognized and widely used feature extraction technique in image classification \cite{dalal2005histograms}. The pixel-wise histograms of gradient directions are computed to generate the HOG descriptor. Consequently, the HOG descriptor utilizes this distribution of local gradient for the accurate detection of complex shapes as well as to object deformation, e.g. facial expression and muscle movement \cite{chen2014facial}. The HOG features are invariant to geometric transformation, light conditions, and color variations. These features may assist the lateralized system to accurately classify images based on facial expressions.

\subsubsection{Adversarial Attacks}
An adversarial attack is a technique to subtly modify an image, by adding a small (imperceptible to a human) perturbation, such that a classifier can accurately classify the original image but misclassify the modified (adversarial) image. Adversarial attacks generate uncertainty, noise, irrelevant, and redundant data. Thus they provide an ideal test-bed in which to evaluate robustness of an AI system. For this work, two well-known strategies are applied to generate adversarial attacks, i.e. fast gradient sign method (FGSM) and iterative adversarial method \cite{goodfellow2014explaining,kurakin2016adversariale}. 

\subsubsection{Deep Learning}
\label{DeepL}
A methodology of applying multi-layer artificial neural networks to extract useful patterns and higher-level features from raw data is called deep learning \cite{deng2014deep}. Deep learning is a connectionist approach to create an AI system. Deep learning systems transform each lower layer representation into a higher layer of abstract representation. A credit assignment path is a chain of information transformation between all layers of the network, i.e. from the input layer to the output layer. At the start of the learning process, a deep learning model considers all features homogeneously. As the learning proceeds, it starts placing the features optimally at different levels such that they improve performance \cite{schmidhuber2015deep, bengio2013representation}. In contrast, the lateralized approach enables the AI system to simultaneously consider the features of the input signal at different levels of abstraction, i.e. constituent level and holistic level. This empowers the lateralized system to exhibit robustness against noise, corrupt, and irrelevant data.

\subsubsection{Learning Classifier Systems}
\label{LCS}
Learning classifier systems (LCSs) is a methodology of developing rule-based solutions for complex problems. LCS is a symbolic/sub-symbolic approach to create an AI system. LCSs have the ability to learn patterns in distinct areas of the search space (niches) by applying discovery algorithms and learning components \cite{urbanowicz2009learning, siddique2022learning}. Holland and Reitman created the first LCS and named it ``Cognitive System One'' \cite{holland1998cognitive}, in recognition of its Cognitive Neuroscience foundation. An LCS based artificial agent learns by taking the appropriate actions that maximize the reward from the environment. As learning proceeds, the LCS starts to evolve context-dependent rules which provide knowledge of the environment. LCSs apply this knowledge in a piecewise manner to make predictions \cite{urbanowicz2009learning}.

\subsection{Lateralized System}
Advanced deep learning based AI systems can efficiently and precisely solve the majority of practical visual classification tasks in computer vision problems \cite{deng2014deep}. These systems generate a homogeneous knowledge representation to learn a linearly separable relationship between features and target. However, conventional AI systems are highly vulnerable to adversarial attacks due to their reliance on homogeneous knowledge representation \cite{akhtar2018threat,madry2017towards}. These systems can easily be fooled by a small perturbation (imperceptible to a human) to an image and a single targeted pattern can disrupt the performance accuracy \cite{moosavi2016deepfool,su2019one}. Moreover, conventional AI systems encourage linear behavior to boost the learning process. However, the majority of adversarial attacks exploit this hallmark to fool these systems \cite{goodfellow2014explaining,madry2017towards,kurakin2016adversariale}. Instead of generating a homogeneous knowledge representation to learn this type of linearly separable relationship , the lateralized AI system automatically identifies and learns such relationships at different levels of abstraction, i.e. at constituent level and holistic level \cite{siddique2020lateralized}. Consequently, the lateralized AI system could only be fooled if an adversarial attack successfully disturbs all those relationships. 

\subsection{Method}
\begin{figure}[h]%[h]
	\begin{center}
		\includegraphics [scale=0.41]{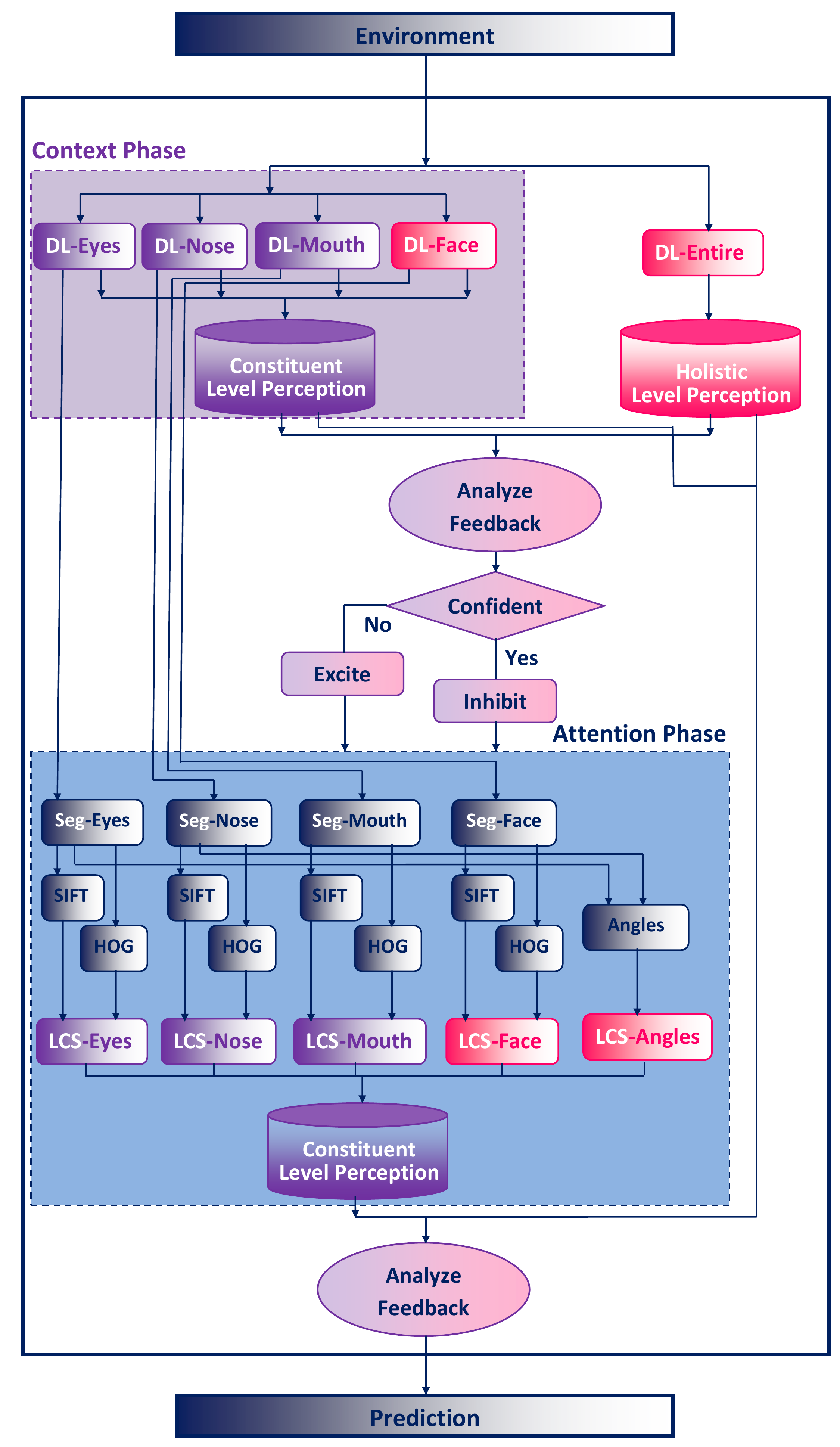}
		%\vspace{-2.5em}
		\caption{A schematic depiction of the strategies developed to achieve cognitive inspired functionality in the lateralized system.}
		\label{FlwChrt4CV}
		%\vspace{-1.5em}
	\end{center}
\end{figure}

The state-of-the-art deep learning (DL) models and LCSs are used to develop a lateralized AI system for computer vision problems. The overall classification process implemented by the lateralized system is divided into two main phases: the context phase and the attention phase. The novel system identifies, generates, and utilizes constituent BBKs and holistic BBKs at each phase. These BBKs (features) are extracted from deep networks and LCSs in the context phase and attention phase, respectively. A schematic illustration of the lateralized AI system developed for computer vision problems is shown in Figure \ref{FlwChrt4CV}.

The \textit{context phase} consists of five DL models. Three of the DL models are used to generate predictions about the individual parts, i.e. constituent level predictions about mouth, nose, and eyes. The remaining two DL models are used to generate predictions about the big picture, i.e. holistic level predictions about the face and overall image. The prediction value is the probability to be associated with a candidate class. This value is normalized between $0$ and $1$. Moreover, a negative or a positive sign is assigned to each prediction value for the cat or dog class, respectively. Subsequently, constituent level perception (CLP) is computed by adding all the constituent level predictions; whereas, holistic level perception (HLP) is computed by adding all the holistic level predictions. If both the CLP and HLP have the same signs then they support each other's perception, and the system can confidently make the final prediction. The system generates an \textit{inhibit} signal to stop further processing in the attention phase. However, if the CLP and HLP have opposite signs then they oppose each other's perception. Consequently, the system is not confident to make the final prediction. In this case, the system generates an \textit{excite} signal to continue further processing in the attention phase.

The \textit{attention phase} consists of five LCSs. Three of the LCSs are used to generate predictions about the individual parts, i.e. constituent level predictions about mouth, nose, and eyes. The remaining two LCSs are used to generate the prediction about the big picture, i.e. holistic level predictions about the face and the angles of a triangle consisting of two eyes and nose vertices (big picture). Generally, a cat's eyes are relatively close to its nose, whereas, a dog's eyes are relatively far from its nose. An imaginary triangle with eyes and nose as vertices is an important distinguishing feature between cats and dogs. This feature is used to get a holistic level prediction.

The same sensory input (problem instance) is presented to the context phase and attention phase. Both the phases start processing in parallel. The attention phase stops its processing immediately on receiving an inhibit signal from the context phase. The attention phase segments individual parts and face of the given image by utilizing the predictions generated by DL models during the context phase. Two types of features are computed from those segmented images, i.e. HOG \cite{dalal2005histograms} and SIFT \cite{lowe2004distinctive}. Subsequently, the LCSs generate constituent and holistic level predictions by utilizing these features. These prediction values are normalized between $0$ and $1$. A negative or a positive sign is assigned to each prediction value to indicate the cat or dog class, respectively. Subsequently, the CLP is computed by adding all the constituent level predictions; whereas, the HLP is computed by adding all the holistic level predictions. Finally, the CLP and HLP values of both phases are added. The system makes the final prediction as a dog if the sign of the final vote is positive, otherwise as a cat. 

\subsection{Experiments}
\begin{figure}%[h]
	\subfloat[\label{ExpCatAnat}]{\includegraphics[scale=0.28]{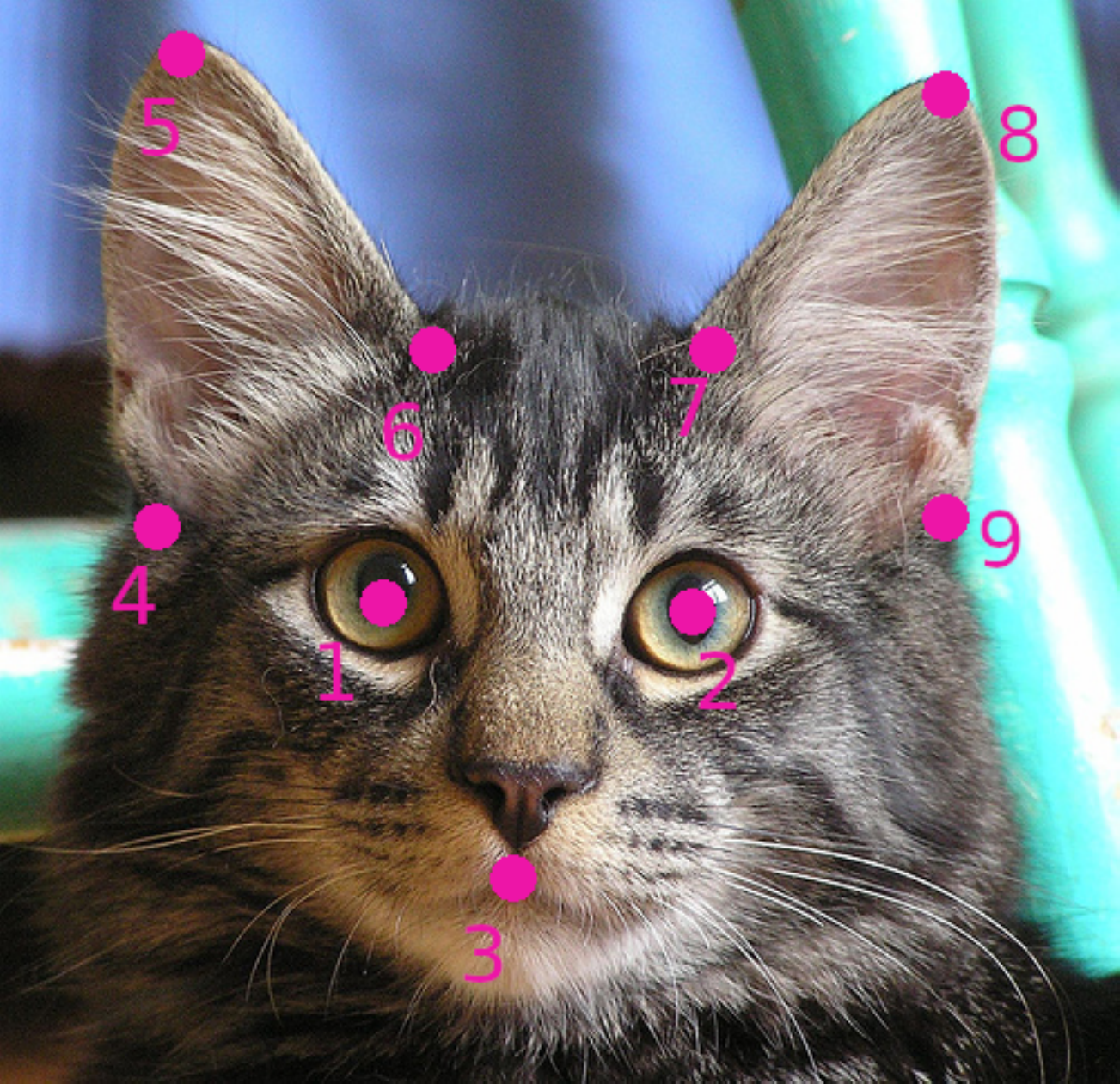} }\hfill
	\subfloat[\label{ExpDogAnat}]{\includegraphics[scale=0.33]{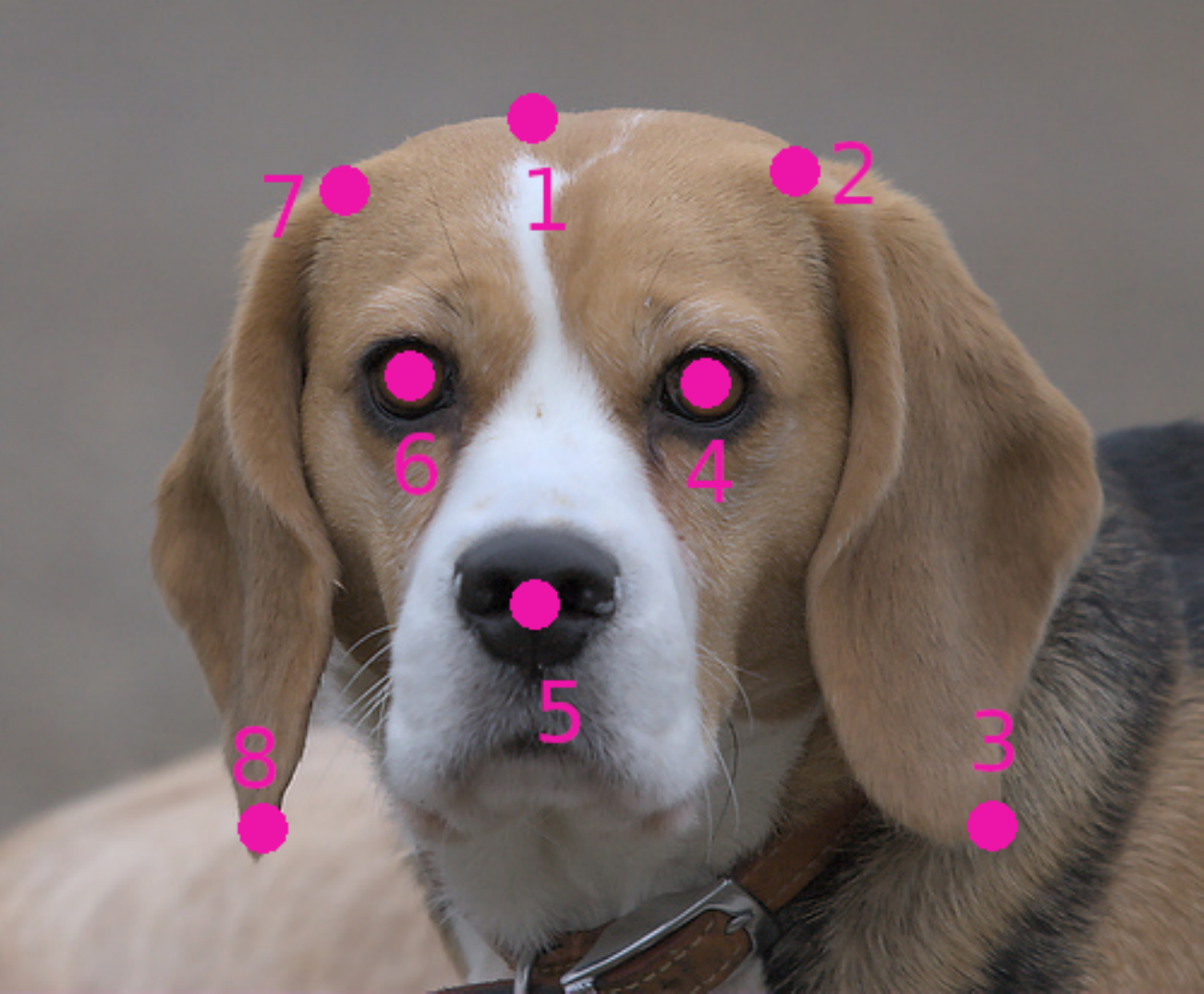}}
	\caption{Sample cat and dog images with $9$ and $8$ points annotation of the head, respectively.}
	\label{SampleImgs}
\end{figure}

In order to use in a lateralized system, the computer vision problems need to have ground-truth information about both the constituent level (parts) and holistic level (whole) representations. The cat vs dog classification experiments were conducted on well-known publicly available data sets \cite{kaggleCats, dlibDogs}. These datasets have ground-truth information about the parts (nose, eyes, ear, mouth, head, face) and the overall image. The cat dataset was taken from Kaggle competition \cite{kaggleCats}. It includes more than $9000$ cat images along with ground-truth files. Each image contains $9$ points annotation of the head, i.e. (1) Left Eye, (2) Right Eye, (3) Mouth, (4) Left Ear-1, (5) Left Ear-2, (6) Left Ear-3, (7) Right Ear-1, (8) Right Ear-2, (9) Right Ear-3 (see Fig. \ref{ExpCatAnat}). 

The dog dataset was taken from dlib (C++ library for ML) \cite{dlibDogs}, which is a modified copy (modified missed annotations and loose BBoxs) of the data used by Liu et al. \cite{liu2012dog}. It includes more than $8000$ dog images along with the ground-truth information. Each image contains $8$ points annotation of the head, i.e. (1)head top, (2) left ear base, (3) left ear tip, (4) left eye, (5) nose, (6) right ear base, (7) right ear tip, (8) right eye (see Fig. \ref{ExpDogAnat}).

The data sets were randomly divided into $80\%$ training-images and $20\%$ test-images. The novel system is trained by using the original train-images only. Subsequently, the adversarial attacks are applied only to the test-images. Two variants of the FGSM adversarial attacks and two variants of iterative (Itr) adversarial attacks are applied to generate adversarial images, i.e. strong attack FGSM-S, medium-level attack FGSM-M, strong attack Itr-S, and medium-level attack Itr-M. Four state-of-the-art deep networks were trained and tested on the same set of training-images, test-images, and adversarial-images i.e. AlexNet \cite{krizhevsky2012imagenet}, ResNet \cite{he2016deep}, SqueezeNet \cite{iandola2016squeezenet}, and VGG \cite{simonyan2014very}. The experimental results of the novel lateralized system (LateralSys) were compared with the results from these four deep networks.

\begin{table}[!t]
	\caption{Classification Accuracy \\ 
            \scriptsize Highest Accuracy is in bold.}
    %\vspace{-0.80em}
	\label{ExpResTable}
	\resizebox{\columnwidth}{!}{%
	\begin{tabular}{|c|c|c|c|c|c|}
        \hline
        \textbf{}         & \textbf{VGG} & \textbf{SqueezeNet} & \textbf{AlexNet} & \textbf{ResNet} & \textbf{LateralSys} \\ \hline
        \textbf{OrigImgs} & 99.37        & 98.35               & 97.24            & 98.64           & \textbf{99.80}               \\ \hline
        \textbf{FGSM-M}   & 94.68        & 85.91               & 90.25            & 89.30           & \textbf{98.34}               \\ \hline
        \textbf{FGSM-S}   & 70.38        & 55.22               & 63.00            & 77.61           & \textbf{81.06}               \\ \hline
        \textbf{Itr-M}    & 97.44        & 95.81               & 94.72            & 88.24           & \textbf{99.59}               \\ \hline
        \textbf{Itr-S}    & 96.10        & 93.91               & 93.66            & 83.41           & \textbf{99.30}               \\ \hline
        \end{tabular}
	}%end of resizebox
	%\vspace{-1.20em}
\end{table}

The experimental results show that the novel lateralized system outperformed the state-of-the-art deep networks in all experiments, see Table \ref{ExpResTable}. The novel system managed to maintain the classification accuracy (loss is less than $1\%$) and exhibits strong robustness against the majority of the adversarial attacks, i.e. Itr-S, Itr-M, and FGSM-M, whereas all other deep networks lost classification accuracy by $5\%$ to $26\%$. Moreover, the classification accuracy of the novel system decreased to $81.06\%$ against the FGSM-S attack. The FGSM-S is a very strong adversarial attack. Consequently, this attack destroys image content very badly. Despite this disruption, the novel system performed better than all other deep networks.

The novel LateralSys considers the given input image at different levels of abstraction. This functionality adds extra computational cost. The context phase of the novel system consists of five deep networks. All these deep networks work in parallel. While this phase utilizes five times more computations than the conventional systems, due to parallel processing the execution time is the same. Moreover, the attention phase consists of four LCSs that also work in parallel. These systems are required to do more work only for complex situations. This attention phase doubles the execution time. However, the computational cost in this phase, is minimized by utilizing excite and inhibit signals. The overall extra cost of the novel system is justified by the robustness against adversarial attacks. An adversarial attack needs to successfully challenge the constituents and holistic components of an image to fool the novel LateralSys.

The interpretation of decisions made by the lateralized system explains the gain in performance accuracy and robustness against adversarial attacks. To illustrate, four examples from cat images and four examples from dog images are selected. These examples are the complex images where the systems' modules are at odds with each other but the final decision made by the novel system is correct. The whole process is explained in the following three scenarios:

\begin{figure}
	\begin{center}
		\includegraphics [scale=0.5]{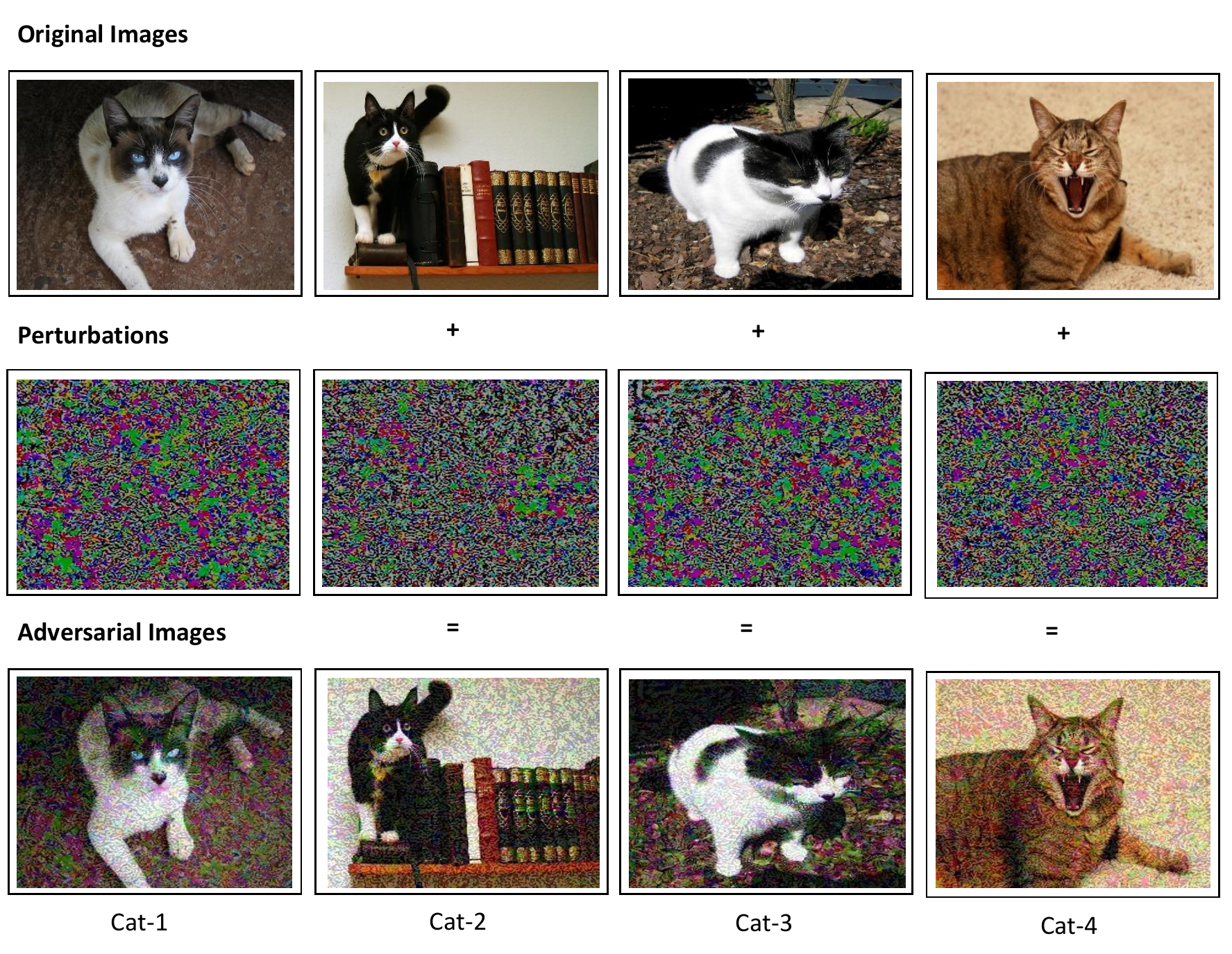}
		%\vspace{-2.0em}
		\caption{Example cat images. Original images are in the first row. The relevant perturbations are in the second row. The resultant adversarial images are in the third row.}
		\label{CatImgs4IntpExp}
		%\vspace{-0.6em}
	\end{center}
	%\vspace{-1.50em}
\end{figure}

\subsubsection{Scenario-1}
\begin{figure}
	\begin{center}
		\includegraphics [scale=0.49]{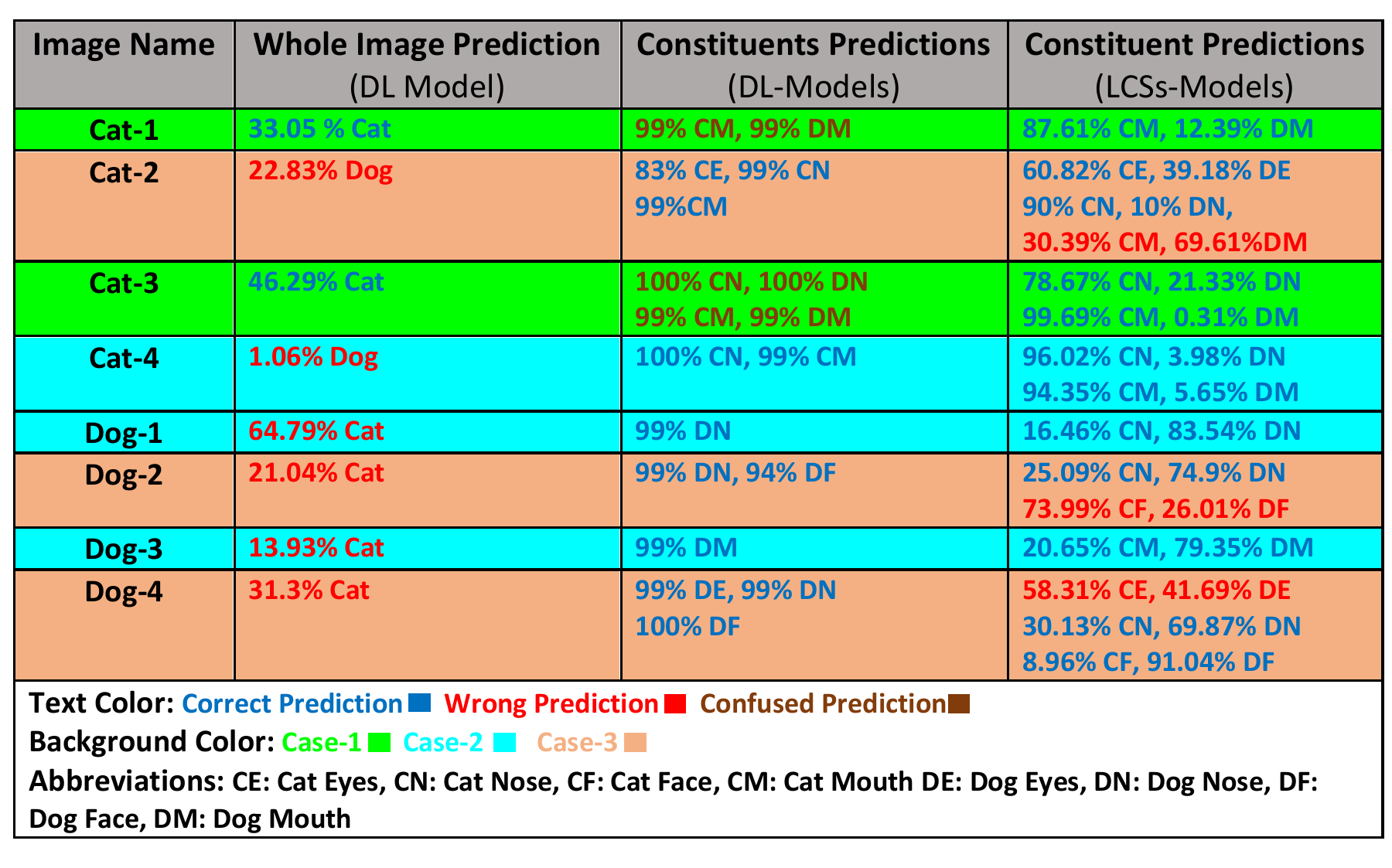}
		%\vspace{-2.0em}
		\caption{Interpretation of decision-making process adopted by the lateralized system to classify adversarial images.}
		\label{InterpRes}
		%\vspace{-0.6em}
	\end{center}
	%\vspace{-1.50em}
\end{figure}

In this case, the deep networks of the context phase are at odds with each other such that HLP is correct with low confidence, and CLP is confused due to opposite class predictions made by the constituent deep networks. Consequently, the system generates an excite signal to do further analysis at the attention phase. Here LCSs models make correct predictions with high confidence that align with the HLP from the context phase. It assists the novel system to generate correct final prediction with high confidence. 
For example, the deep networks at the context phase predicted that the Cat-1 image has $99.00\%$ dog mouth, $99.00\%$ cat mouth; but overall it looks, $33.05\%$, like a cat. The constituent level deep networks are confused whether the image is a cat or a dog. Consequently, the system generated an excite signal. The LCSs at the attention phase predicted that the Cat-1 image has a $12.39\%$ dog mouth and $87.61\%$ cat mouth. The constituent level LCSs are very clear that it is a cat. These predictions aligned with the HLP at the context phase and assisted the novel system to make a correct final prediction (i.e. cat) with high confidence, see Figs. \ref{CatImgs4IntpExp} and \ref{InterpRes}.

Similarly, the deep networks at the context phase predicted that the Cat-3 image has a $100.00\%$ dog nose, $100.00\%$ cat nose, $99.00\%$ dog mouth, and $99.00\%$ cat mouth; but overall it looks, $46.29\%$, like a cat. The constituent level deep networks are confused whether the image is a cat or a dog. Consequently, the system generated an excite signal. The LCSs at attention phase predicted that the Cat-3 image has a $21.33\%$ dog nose, $78.67\%$ cat nose, $0.31\%$ dog mouth, and $99.69\%$ cat mouth. The constituent level LCSs are very clear that it is a cat. These predictions aligned with the HLP at the context phase and assisted the novel system to make a correct final prediction (i.e. cat) with high confidence, see Figs. \ref{CatImgs4IntpExp} and \ref{InterpRes}

\subsubsection{Scenario-2}
In this case, the deep networks of the context phase are at odds with each other such that constituent level deep networks generate correct predictions, whereas the holistic level deep networks make wrong predictions. Consequently, the system generates an excite signal to do further analysis at the attention phase. Here LCSs models make correct predictions with high confidence that align with the CLP from the context phase. It assists the novel system to generate correct final prediction with high confidence. 

For example, the deep networks at the context phase predicted that the Cat-4 image has a $100.00\%$ cat nose, $99.00\%$ cat mouth; but overall it looks, $1.06\%$, like a dog. The holistic level deep networks made an opposite prediction with low confidence. Consequently, the system generated an excite signal. The LCSs at attention phase predicted that the Cat-4 image has a $3.98\%$ dog nose, $96.02\%$ cat nose, $5.65\%$ dog mouth, and $94.35\%$ cat mouth. The constituent level LCSs are very clear that it is a cat. These predictions aligned with the CLP at the context phase and assisted the novel system to make a correct final prediction (i.e. cat) with high confidence, see Figs. \ref{CatImgs4IntpExp} and \ref{InterpRes}.

\begin{figure}
	\begin{center}
		\includegraphics [scale=0.5]{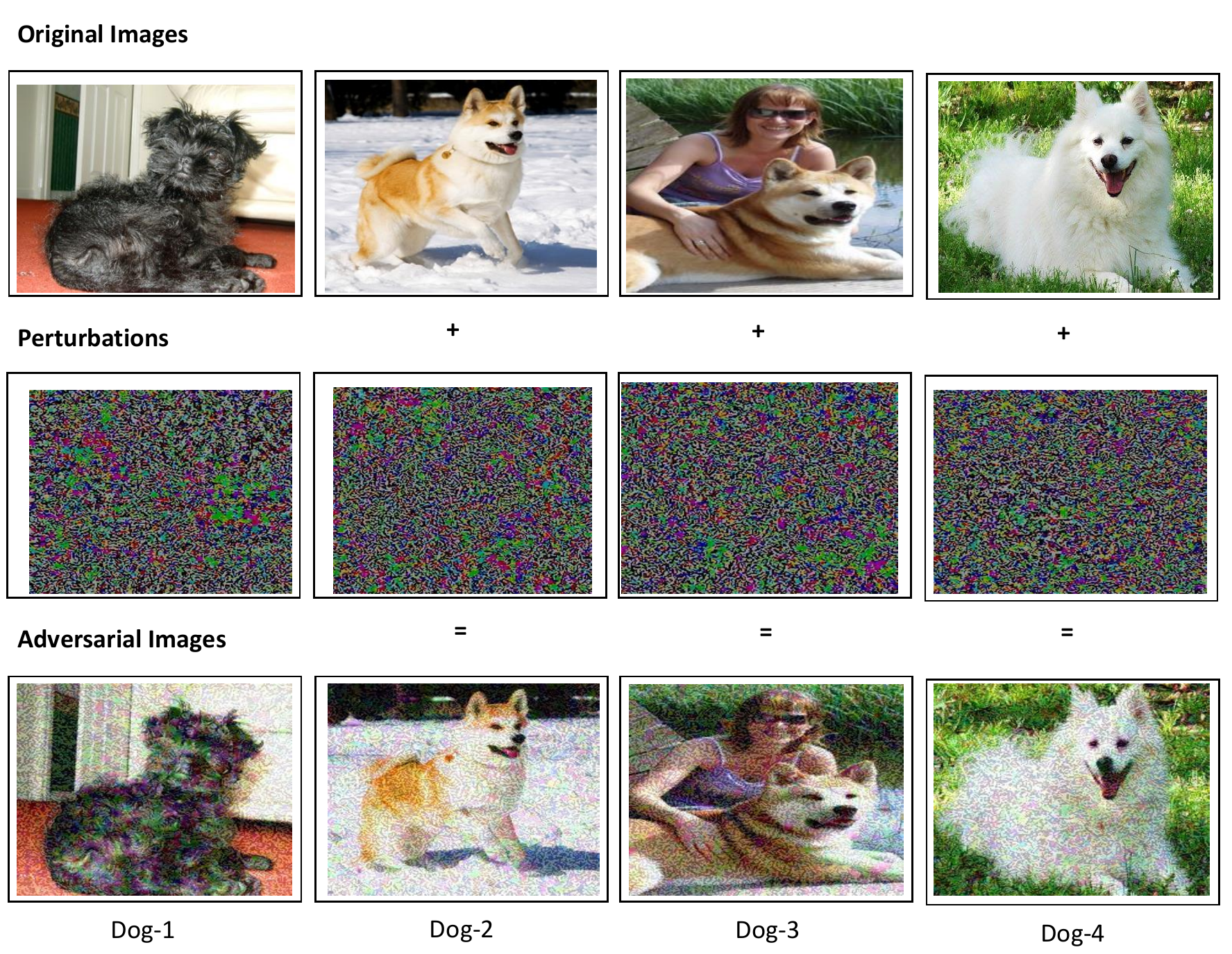}
		%\vspace{-2.0em}
		\caption{Example dog images. Original images are in the first row. The relevant perturbations are in the second row. The resultant adversarial images are in the third row.}
		\label{DogImgs4IntpExp}
		%\vspace{-0.6em}
	\end{center}
	%\vspace{-1.50em}
\end{figure}

Similarly, the deep networks at the context phase predicted that the Dog-3 image has a $99.00\%$ dog mouth; but overall it looks, $13.93\%$, like a cat. The constituent level deep networks and holistic level deep networks are at odds with each other. Consequently, the system generated an excite signal. The LCSs at the attention phase predicted that the Dog-3 image has a $79.35\%$ dog mouth and $20.65\%$ cat mouth. The constituent level LCSs are very clear that it is a dog. These predictions aligned with the HLP at the context phase and assisted the novel system to make a correct final prediction (i.e. dog) with high confidence, see Figs. \ref{DogImgs4IntpExp} and \ref{InterpRes}.

\subsubsection{Scenario-3}
In this case, the deep networks of the context phase are at odds with each other such that HLP is wrong and CLP is partially confused due to weak class predictions made by the constituent deep networks. Consequently, the system generates an excite signal to do further analysis at the attention phase. Here LCSs models make correct predictions with high confidence that aligned with the CLP from the context phase. It assists the novel system to generate correct final prediction with high confidence. 

For example, the deep networks at the context phase predicted that the Cat-2 image has a $99.00\%$ cat nose, $83.00\%$ cat eyes, and $99.00\%$ cat mouth; but overall it looks, $22.83\%$, like a dog. The constituent level deep networks and holistic level deep networks are at odds with each other. Consequently, the system generated an excite signal. The LCSs at attention phase predicted that the Cat-2 image has a $39.18\%$ dog eyes, $60.82\%$ cat eyes, $10\%$ dog nose,  $90\%$ cat nose, $69.61\%$ dog mouth, and $30.39\%$ cat mouth. The constituent level LCSs are also at odds with each other but the overall vote from LCSs models supports the correct decision. The correct predictions from the context phase and attention phase outweigh the wrong predictions. These predictions assisted the novel system to made a correct final prediction (i.e. cat) with high confidence, see Figs. \ref{CatImgs4IntpExp} and \ref{InterpRes}.

\section{Case Study 2: Heterogeneous AI System for Navigation Problems}
\label{CaseStudy2}
Visual identification tasks are single-step problems in spatial domains. However, the majority of the biological tasks where heterogeneity \footnote{Lateralization can be considered as a special type of heterogeneity.} is applied, are temporal in nature. The majority of the AI systems struggle to capture complex environmental structures in temporal domain. The navigation problems are multi-step path planning problems that provide a virtual environment that simulates real-world navigational problems. The second system was developed to investigate the effectiveness of the lateralized approach in resolving complex navigation problems, where we know that vertebrate brains use multiple levels of representation simultaneously.

Perceptual aliasing is a long-standing problem for artificial agents in applying reinforcement learning (RL) to multi-step navigational tasks \cite{whitehead1991learning,chrisman1992reinforcement,lanzi2000adaptive,butz2002anticipatory,zatuchna2005agentp,krening2019newtonian}. An artificial agent is assumed to perceive its local environment without access to the global world view, which it needs to construct through interaction with the environment. Aliasing occurs when the agent's internal representation confounds the external world states, i.e. the agent's current perception is unable to distinguish environmental states which appear identical but require different actions \cite{whitehead1991learning}. In such a scenario, the reinforcement for the environment instructs the agent to take a specific action in a given state. Unfortunately, when the agent encounters an aliased state, it persists in taking the same action, which will now be reinforced differently. This inconsistency prevents the learning of stable policies, especially for multi-step tasks \cite{crook2003learning}. Perceptual aliasing, therefore, diminishes the effectiveness of reinforcement learning \cite{chrisman1992reinforcement} and hinders its application to real-world problems \cite{suzuki2019hierarchical}. 

\begin{figure}%[H]
	\begin{center}
		\includegraphics [scale=0.59]{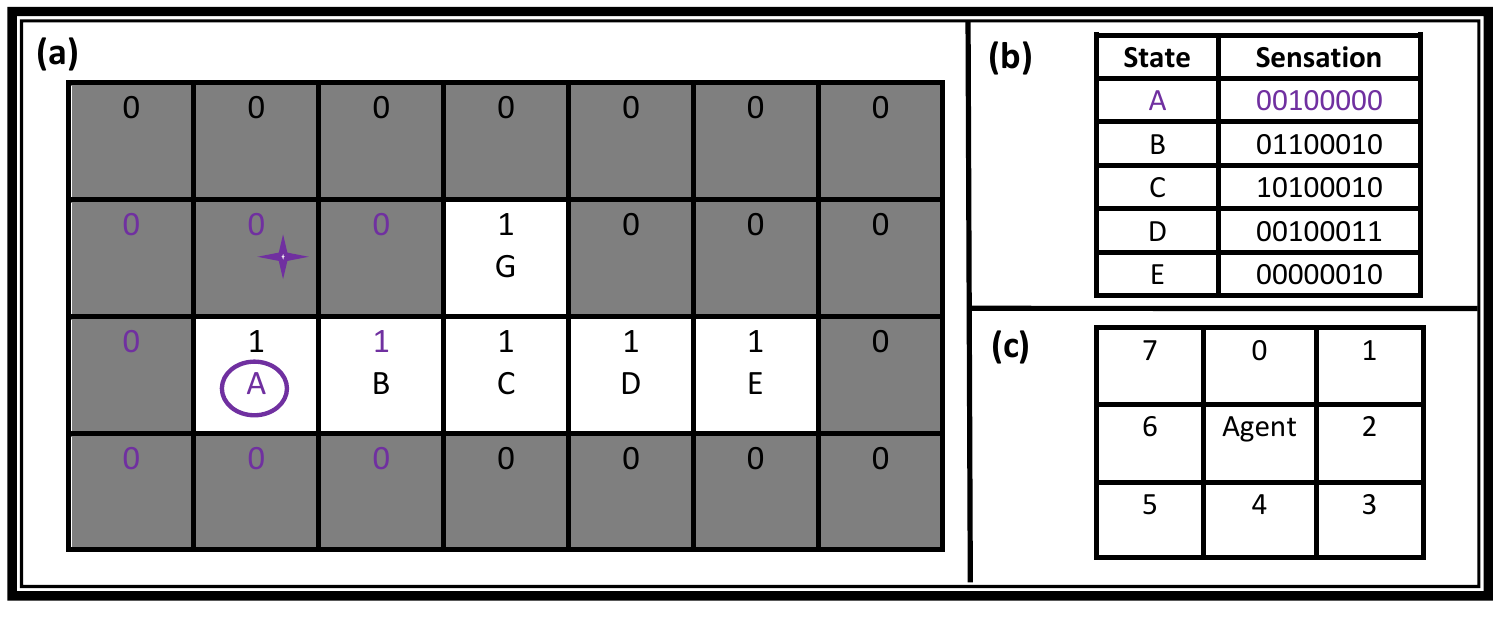}
		%\vspace{-1.3em}
		\caption{(a) A sample maze, here an empty state is represented by $1$ and a blocked state is represented by $0$. (b) States with sensation values, e.g. starting from the north (star), the sensation value of state `A' is $00100000$. (c) An agent with corresponding action values to visit the respective adjacent state. }
		\label{MazeStateAction}
		%\vspace{-2.0em}
	\end{center}
\end{figure}

A navigation problem can be considered as a multi-step path planning problem in an environment (e.g. a maze). Mazes have been commonly used as a testing paradigm in navigation based studies, from cognitive neuroscience to artificial intelligence \cite{orhand2020bacs,alexander2015retrosplenial,alexander2017spatially,nitz2011path,oess2017computational,o1971hippocampus,butz2002anticipatory,butz2006rule,lanzi2000toward,zatuchna2009learning,metivier2002anticipatory}. A maze is a two dimensional rectangular grid, which provides a virtual environment for navigation. A maze provides an ideal environment to evaluate the effectiveness of artificial agents in the temporal domain. Each cell of the two dimensional grid is considered as a state. A state can be empty or blocked. The agent can execute an action and transit to a neighboring empty state but it can not visit a blocked state. For this work, an empty state is represented by $1$ and a blocked state is represented by $0$. There are eight possible actions that an agent can execute to visit an adjacent state. These actions are represented by incremental numbers from $A_{0}$ (starting from the top) to $A_{7}$, as shown in Fig. \ref{MazeStateAction}-c. 

\begin{figure}%[h]%[H]
	\begin{center}
		\includegraphics [scale=0.75]{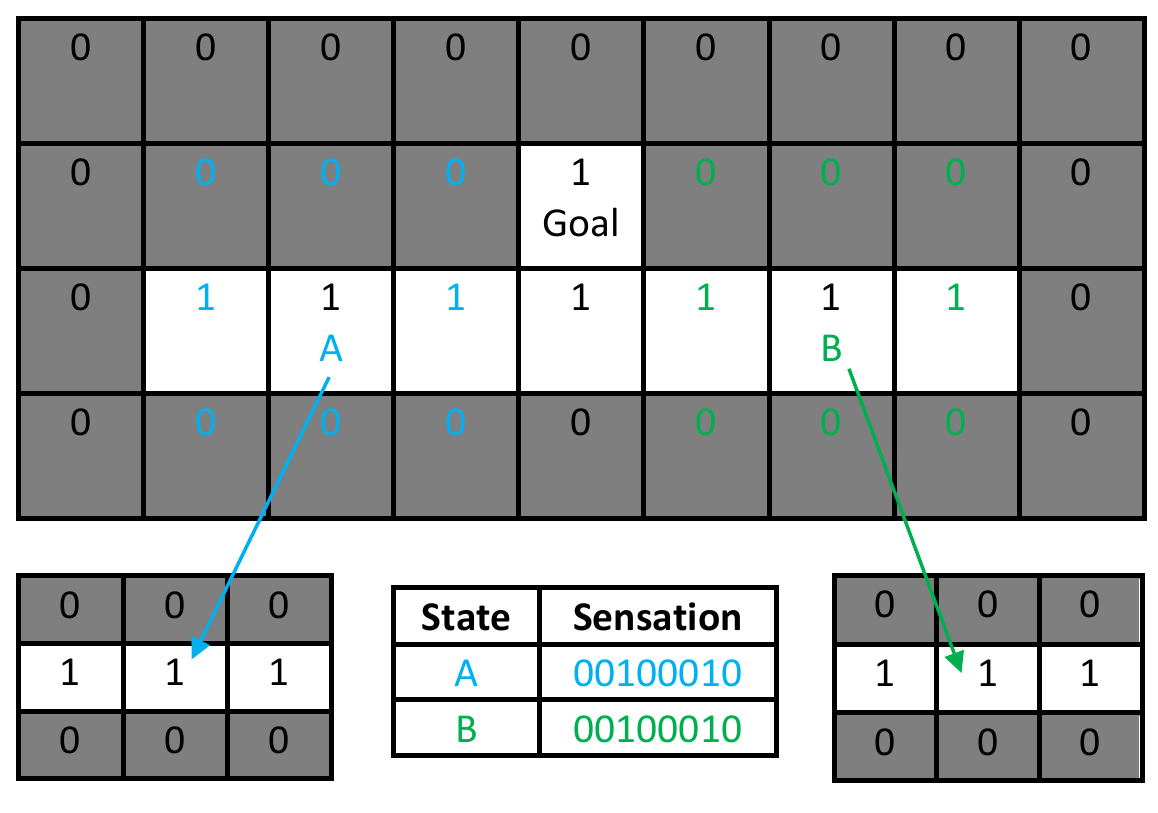}
		%\vspace{-2.0em}
		\caption{A sample maze with two aliased states. }
		\label{non-MarkovMaze}
		%\vspace{-2.0em}
	\end{center}
\end{figure}

An environmental state (input signal) is represented by a binary string which is computed by concatenating the agent's immediate sensation of eight adjacent states. For example, in Fig \ref{MazeStateAction} starting from the north (star), the input signal for state A is ``00100000''. The third bit is `1' because the state on the east side of the state A is empty. A sample maze (without aliased states) and input signal (binary string representation) of each empty state are shown in Figs. \ref{MazeStateAction}-a, \ref{MazeStateAction}-b. Another sample maze environment (with aliased states) is shown in Fig. \ref{non-MarkovMaze}. The states A and B are two aliased states. In these states, the agent's immediate sensation provides the same input signal, i.e. $00100010$. But the agent needs to take different actions to optimally reach the goal state. That is, in-state A, the agent needs to take action `$2$' to move to the right state; whereas, in-state B, the agent needs to take action `6' to move to the left state. Consequently, the agent is facing a perceptual aliasing problem. A large number of approaches have been investigated to handle perceptual aliasing problems \cite{orhand2020bacs,hayashida2017improved,hayashida2014xcs,griffith2013policy,zatuchna2009learning,zatuchna2005agentp,roy2005finding,mitchell2003using,butz2002anticipatory,metivier2002anticipatory,lanzi2000adaptive,stolzmann1999latent,lanzi1998analysis,chrisman1991intelligent,whitehead1991learning,tan1991cost}. These techniques can solve simple maze environments but can not optimally resolve the majority of complex maze environments.

One factor that may have hampered progress in learning in navigation problems is the reliance on a local frame-of-reference (FoR) only, i.e. local viewpoint of the environment based on an agent's immediate perception. Hence the agent could not consider the environment at a higher level of abstraction (big-picture) to uniquely identify aliased states. Consequently, aliased steps in a policy\footnote{A policy, like a route, can be considered as a large pattern prescribing state transitions from a starting state to the goal state.} are stored with the same weight, as non-aliased steps. We assert that an aliased state is a small pattern that gets repeated in an environment, which makes it difficult to identify where it occurs locally.

These patterns can be combined with other patterns (aliased or not) to form higher level patterns and so forth. Eventually, each pattern, either in itself or part of a higher level pattern, is unique. Thus, complex environments entail patterns that form hierarchical patterns, such as multiple aliased states at different positions in the environment. These aliased states can be identified uniquely (i.e. turned to non-aliased states) by considering an environment at different levels of abstraction. Conventional RL systems struggle to capture such complex structures.

Heterogeneous features based AI agents efficiently learn those complex patterns by utilizing BBKs at constituent and holistic levels simultaneously to represent the local viewpoint and world viewpoint (complete map or big-picture) of the environment, respectively \cite{siddique2021frames}. Biological intelligence supports this type of heterogeneity. For example, egocentric frame-of-reference (FoR) is utilized to obtain the local viewpoint; whereas allocentric and routecentric FoRs are utilized to capture the world viewpoint of the environment. FoRs enable vertebrate (and many invertebrates) brains to process the same information at multiple levels of abstraction \cite{corballis2017evolution,krichmar2008neuromodulatory}. The novel system incorporates these principles to handle aliased states in navigation problems.

\subsection{Method}
\begin{figure}[h]%[H]
	\begin{center}
		\includegraphics [scale=0.78]{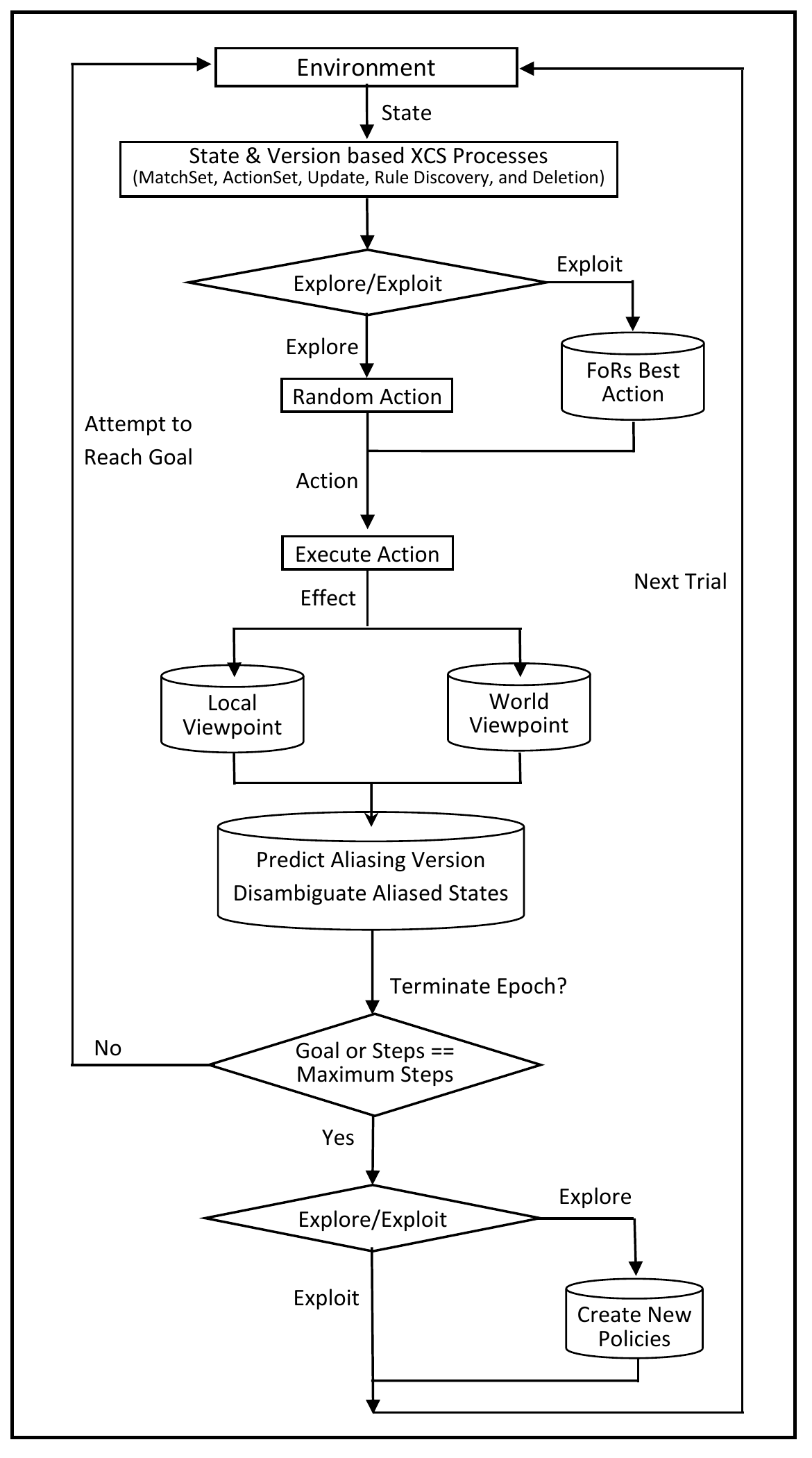}
		%\vspace{-1.3em}
		\caption{A schematic depiction of the overall strategy developed to achieve cognitive inspired functionality. The frame-of-reference (FoR) based system utilizes local viewpoint (constituent BBKs) and world viewpoint (holistic BBKs) to disambiguate aliased states.}
		\label{FlwChrt4Navigation}
		%\vspace{-2.0em}
	\end{center}
\end{figure}

LCSs have been commonly used to conduct research on navigation problems \cite{butz2002anticipatory,butz2006rule,lanzi2000toward,metivier2002anticipatory,zatuchna2009learning,siddique2020learning}. The novel system is developed by using the framework of accuracy based LCSs, i.e. Wilson's XCS \cite{wilson1995classifier}. The multi-step classification task is learned by applying the reinforcement learning technique. The general state-action-reward scheme of the novel system is similar to the standard multi-step reinforcement learning scheme \cite{lanzi1998analysis,butz2002anticipatory}. A schematic illustration of the novel AI system developed for navigation problems is shown in Figure \ref{FlwChrt4Navigation}.

\begin{figure}[h]
	\begin{center}
		\includegraphics [scale=0.59]{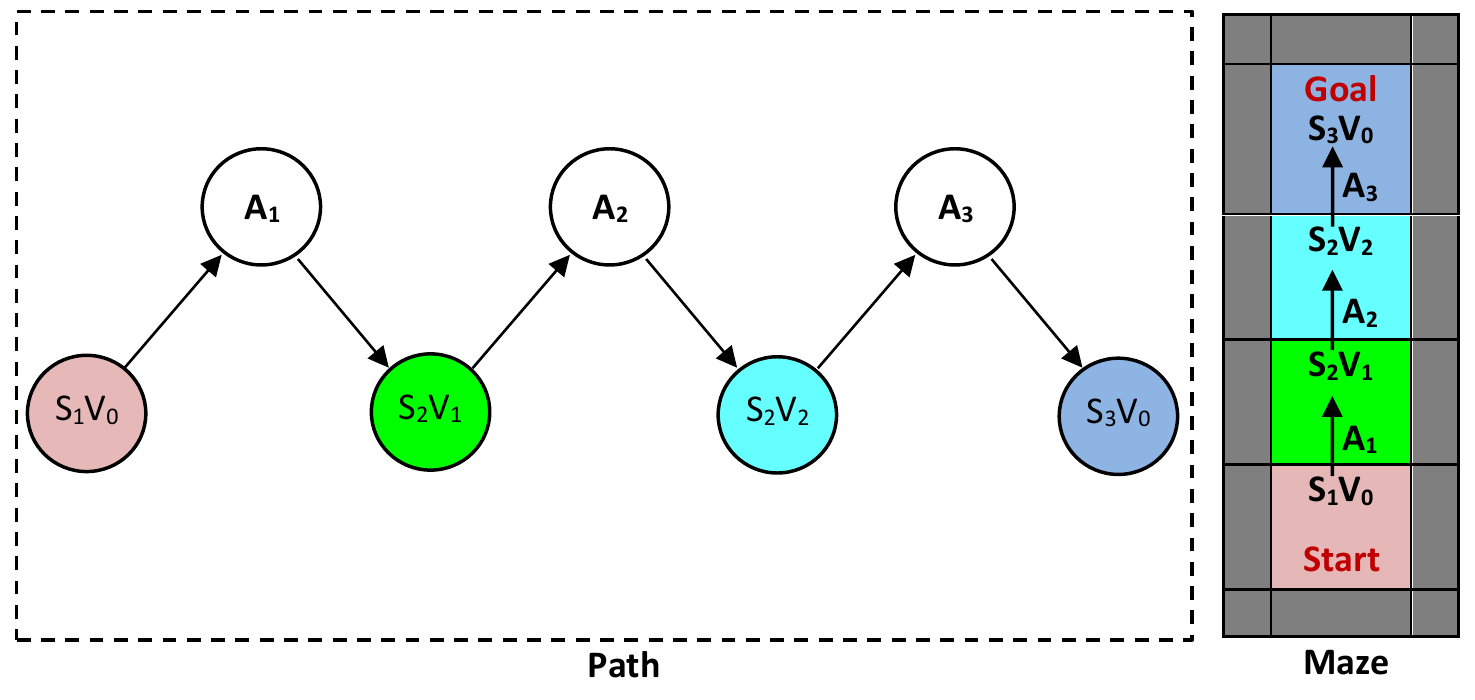}
		%\vspace{-1.50em}
		\caption{A sample maze (right) and corresponding code path (left) when an agent moves from start state to the goal state. States $S_{2}V_{1}$ and $S_{2}V_{2}$ are non-aliased versions of the aliased state $S_{2}$.}
		%\vspace{-0.5em}
		\label{ExampCP}
		%\vspace{-0.60em}
	\end{center}
\end{figure}

Two novel techniques are developed to get local viewpoints (constituents level BBKs) and world viewpoint (holistic level BBKs) of the environment, i.e. code paths and policies. A code path (CP) is a tree like structure that has \textit{state-version} tuple or an \textit{action} in its nodes. Each state of the maze has a unique version that is used to represent multiple existence of same state. The depth of a CP tree is two such that it can have a maximum of seven nodes. The novel system creates/updates the nodes of the respective CP tree while the agent moves from one state ($S_{1}V_{1}$, state-version tuple) to another state ($S_{2}V_{1}$) by executing an action ($A_{1}$). An example CP, when the agent moves form start state to the goal state, is shown in Fig. \ref{ExampCP}. A policy is a holistic level BBK that contains all the CPs required to move from the start state to the goal state. It provides a world viewpoint of the environment. The novel system creates new policies by two methods, i.e. imprinting of utilized CPs and through the evolutionary process.

\begin{table}%[!t]
\centering
	\caption{Performance accuracy of LCS agents \\
	        \scriptsize Lower is better (Optimal performance is in bold)}
	%\vspace{-0.80em}
	\label{ExpResTable}
	%\resizebox{\columnwidth}{!}{%
	%\resizebox{5.5cm}{!}{%
	\begin{tabular}{|c|c|c|c|c|}
        \hline
        \textbf{}          & \textbf{XCSLib} & \textbf{ACS2} & \textbf{AgentP} & \textbf{HetroXCS} \\ \hline
        \textbf{Maze7}     & 31              & 25            & \textbf{4.3}    & \textbf{4.3}        \\ \hline
        \textbf{Littman57} & 12              & 15            & 7               & \textbf{3.8}        \\ \hline
        \textbf{Maze10}    & 40              & 35            & 35              & \textbf{6.5}        \\ \hline
    \end{tabular}
	%}%end of resizebox
	%\vspace{-2.0em}
\end{table}

At the start of the learning process, the agent is randomly placed in the given maze environment. The agent tries to learn the optimal path to reach the goal state by utilizing the least number of steps up to a limit to prevent looping. The agent tries to identify and disambiguate the aliased states during the learning process. For this purpose, the agent utilizes the relevant (i) policies, (ii) multiple CPs, (iii) and multi-step CPs. The agent navigates the environment in two modes, i.e. explore mode and exploit mode. In explore, the agent randomly select an action to take a step. Whereas, in exploit mode, the agent select the best action which can guide the agent to optimally reach the goal state. The new policies are created only during the explore mode. A policy that can lead the agent to the goal state by using minimum steps, discovered so far, is called the best policy. During the exploit mode, the agent activates the best policy. The best action is computed from two methods, i.e. from the action set and the best policy. If these best actions are the same, the agent confidently executes it and marks a flag (`cognate') as true. Otherwise, the agent executes the best action recommended by the active policy and marks the `cognate' flag as false. Subsequently, if the agent is not able to move to the next state or is in a different state expected by the policy, then the agent deactivates the currently active policy, marks it as malign, and activates the new best policy. 

\begin{figure}[h]
	\begin{center}
	    %\vspace{-0.3em}
		\includegraphics [scale=0.52]{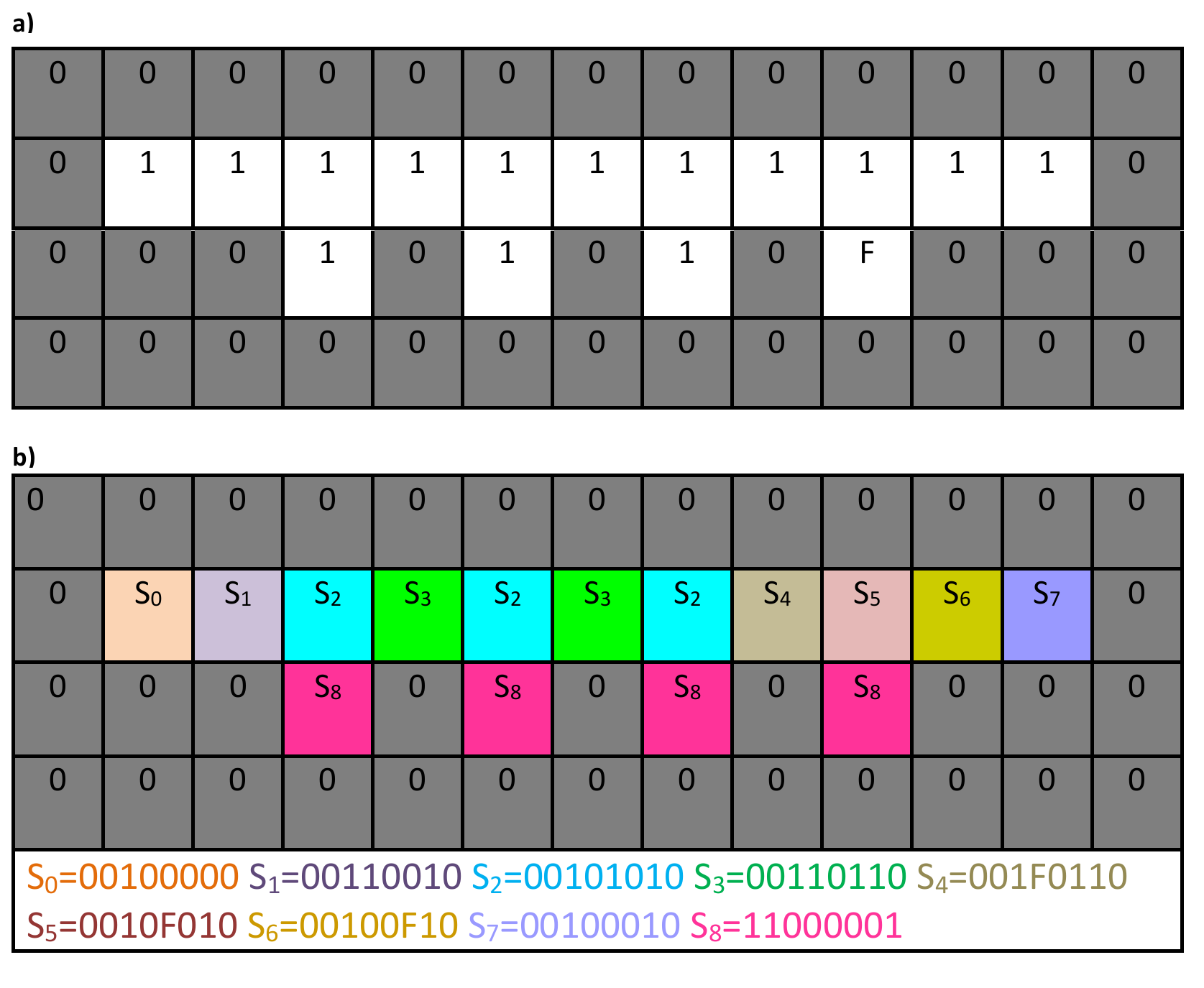}
		%\vspace{-1.80em}
		\caption{(a) Maze57, (b) Aliased states in Maze57. Each color represents a state. Multiple existences of a state indicate aliased states.}
		\label{MapMazeL57}
		%\vspace{-2em}
	\end{center}
\end{figure}

\begin{figure}%[H]
	\begin{center}
	    %\vspace{-0.3em}
		\includegraphics [scale=0.48]{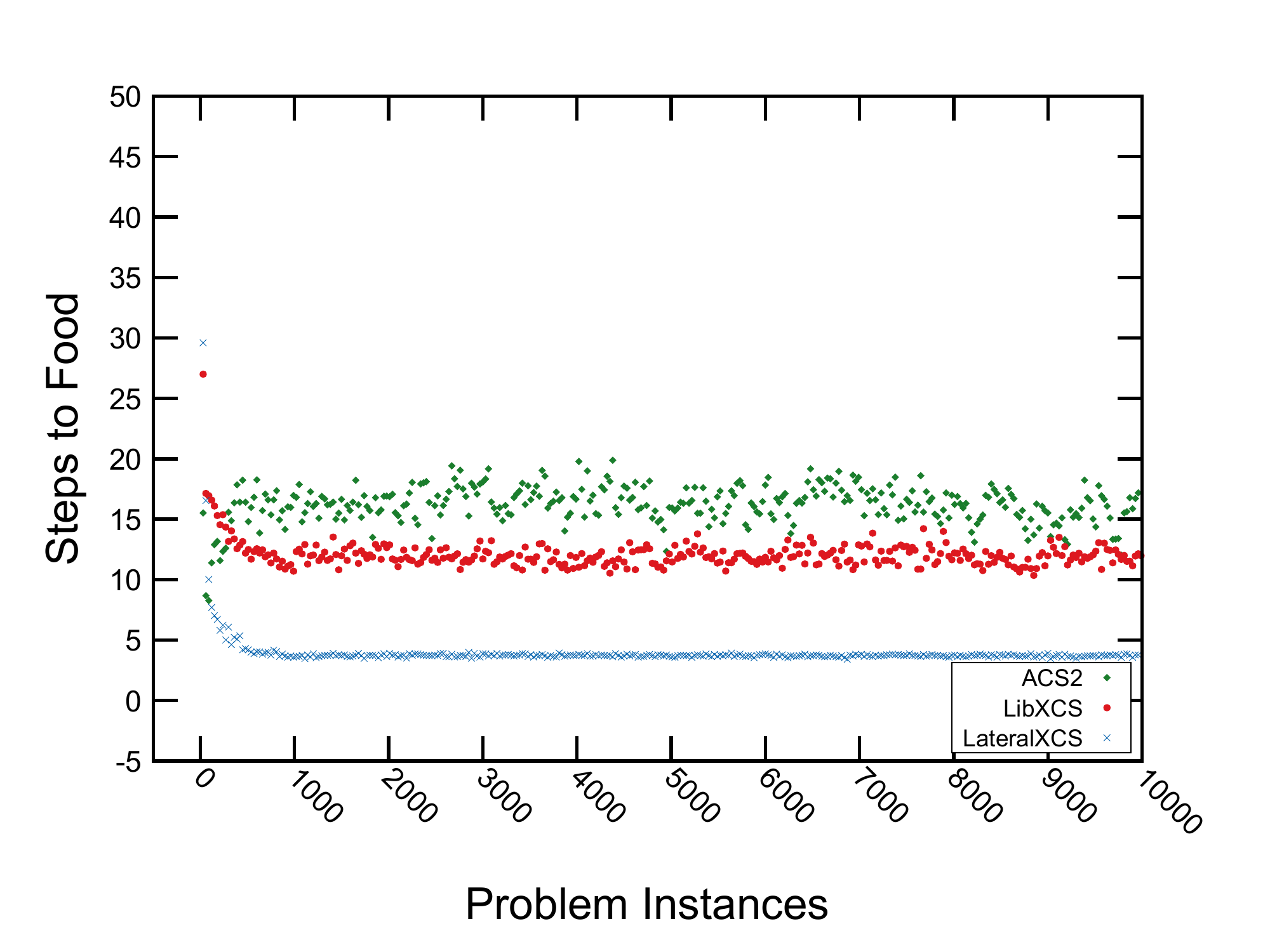}
		%\vspace{-1.80em}
		\caption{Experimental results of Littman57 using ACS2, XCSLib, and heterogeneous features-based XCS (HetroXCS)}
		\label{ExpMazeL57}
		%\vspace{-2em}
	\end{center}
\end{figure}

\subsection{Experiments}
\begin{figure}[h]
	\subfloat[\label{MapMaze10Simple}]{\includegraphics[scale=0.25]{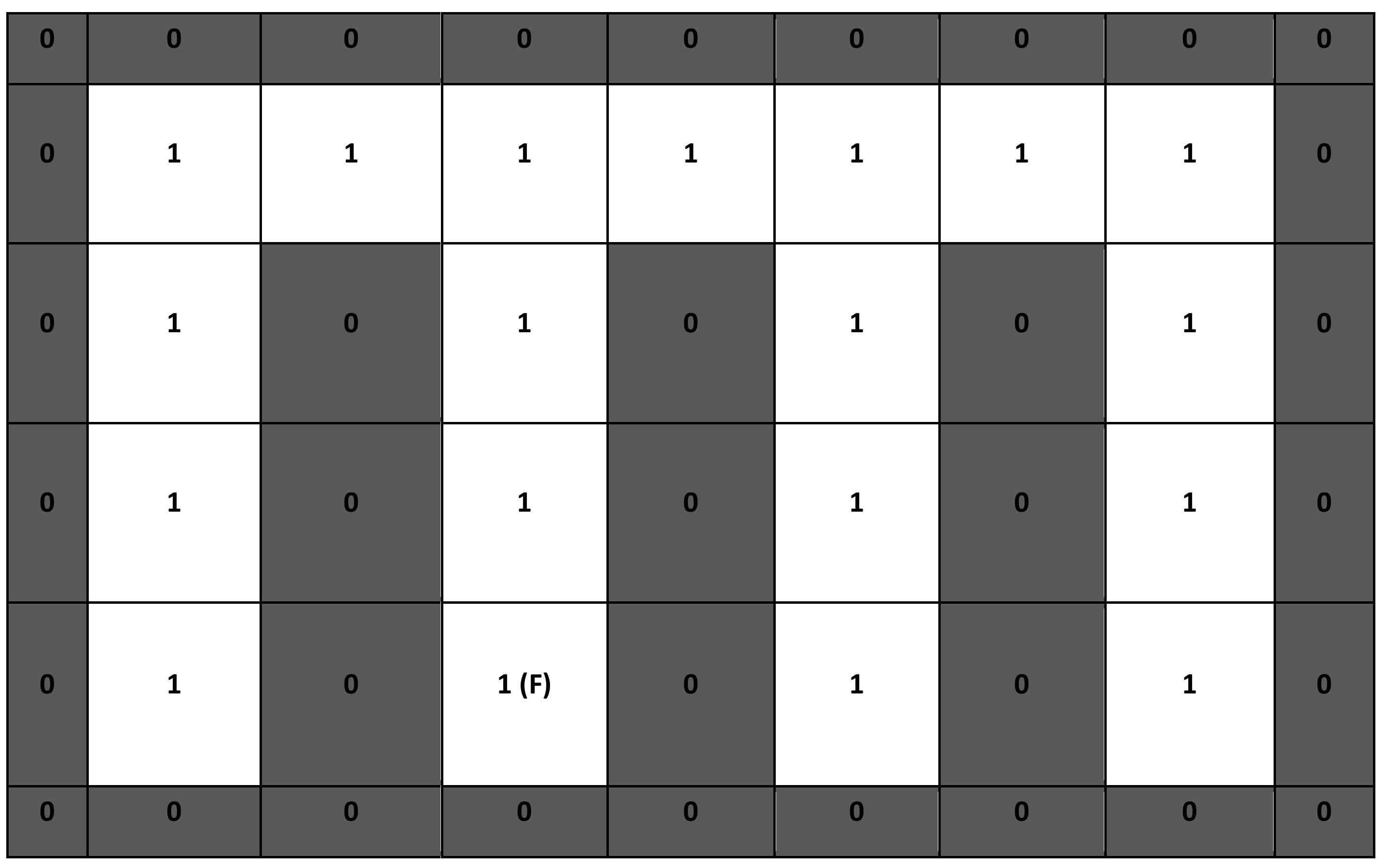}}\hfill
	\subfloat[\label{MapMaze10Aliased}]{\includegraphics[scale=0.25]{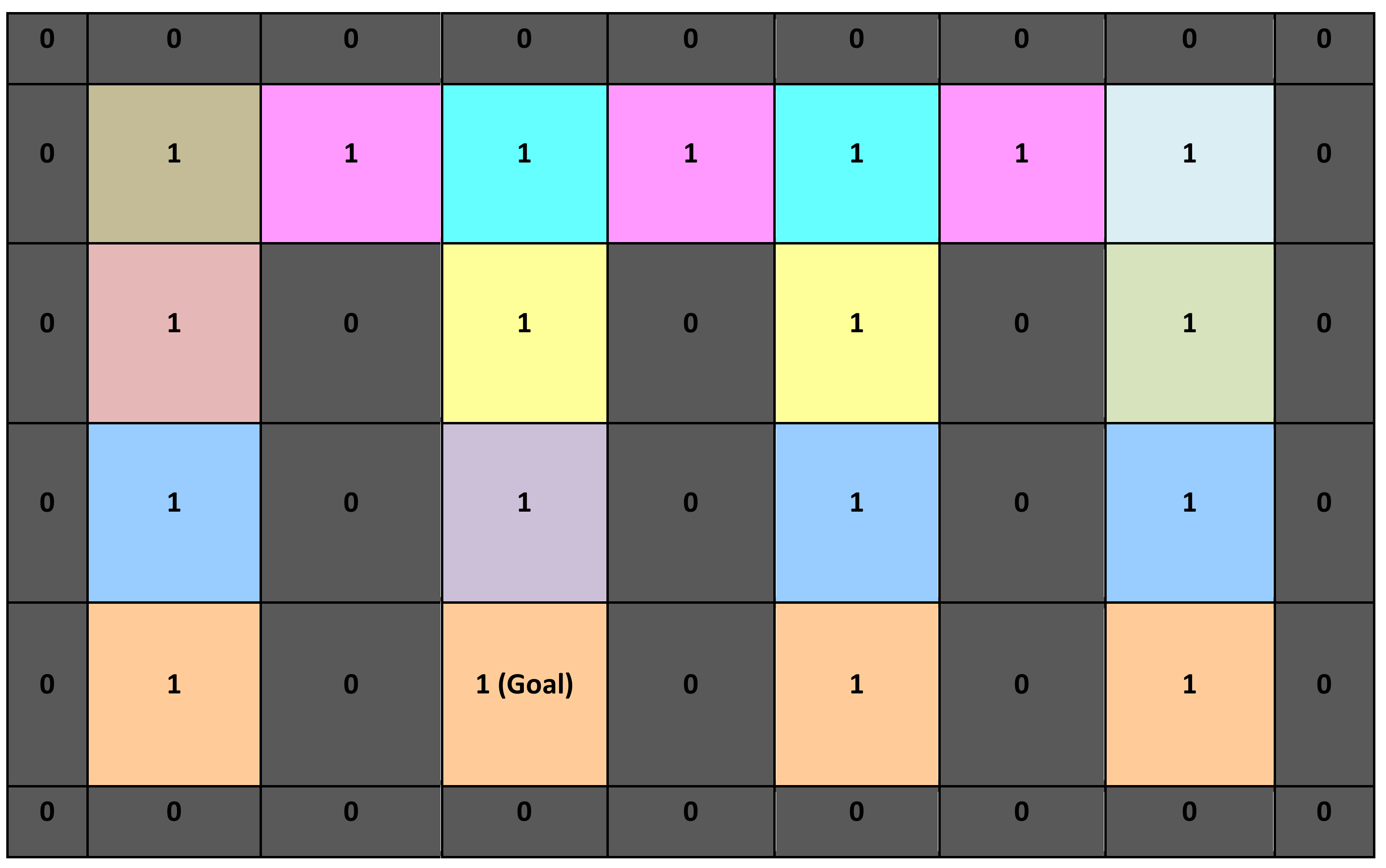}}
	\caption{(a) Maze10, (b) Aliased states in Maze10. Each color represents a state. Multiple existence of a state are aliased states.}
	\label{MapMaze10}
\end{figure}

The navigation experiments are conducted by utilizing three well known complex mazes, i.e. Maze7, Littman57, and Maze10 \cite{zatuchna2009learning}, see Figs. \ref{MapMazeL57}-a and \ref{MapMaze10}. These mazes have clusters (groups) of aliased states that form complex hierarchical patterns. The experimental results of the novel heterogeneous features-based XCS (HetroXCS) are compared with the three state-of-the-art systems, i.e. AgentP (imprinting based system) \cite{zatuchna2009learning}, ACS2 (latent learning based system) \cite{butz2001algorithmic}, and XCSLib (conventional system) \cite{xcslib}. It is noted that we were able to reproduce the experimental results of XCSLib and ACS2, the results of AgentP are taken from the respective studies. 

\begin{figure}%[H]
	\begin{center}
	    %\vspace{-0.3em}
		\includegraphics [scale=0.48]{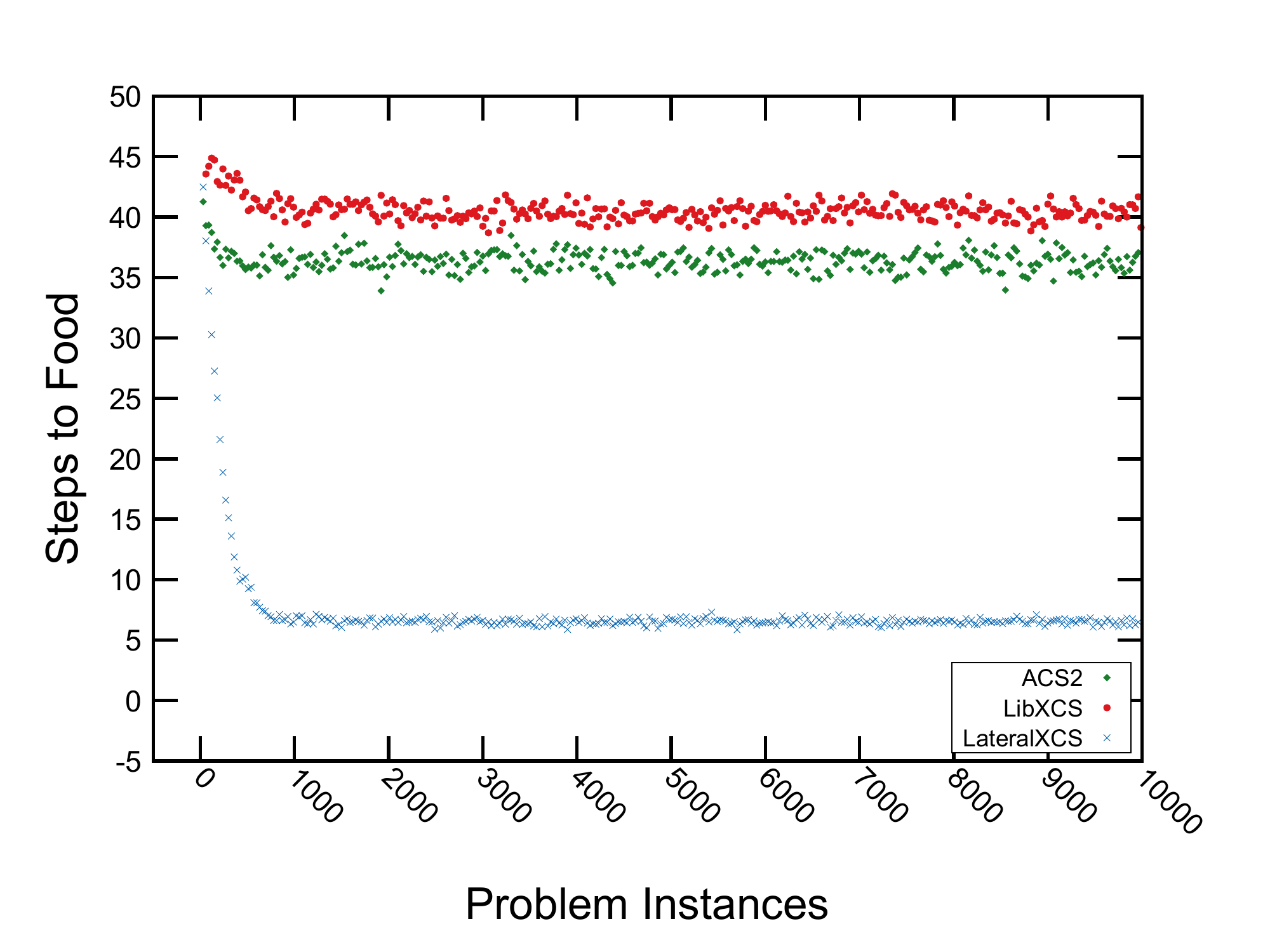}
		%\vspace{-1.80em}
		\caption{Experimental results of Maze10 using ACS2, XCSLib, and HetroXCS}
		\label{ExpMaze10}
		%\vspace{-1.8em}
	\end{center}
\end{figure}

The experimental results of all the systems are presented in Table-\ref{ExpResTable}. Simple and aliased states representations of well-known maze57 are shown in Fig. \ref{MapMazeL57}. The learning pace of different systems for maze Littman57 is compared in Fig. \ref{ExpMazeL57}. Similarly, the complex Maze10 is presented in Fig \ref{MapMaze10}, and learning pace of different systems for Maze10 is shown in Fig. \ref{ExpMaze10}. The experimental results show that all the state-of-the-art systems unable to optimally solve complex mazes Littman57 and Maze10. Whereas, the novel system out-performed other systems and resolved all the mazes by utilizing less number of steps, especially Maze10, that had not been learned to this level previously due to the complex hierarchical patterns of aliased states.

The novel HetroXCS has to utilize multiple viewpoints at different levels of abstraction, depending on the complexity of aliased patterns in the environment. This functionality adds extra computational cost. Although, it is not straightforward to compare the computational cost for different systems due to operating system constraints. These systems could be compared based on the average processing time required for an agent to take a step in an environment. The average single step processing times, computed by using Maze7, for HetroXCS and XCSLib are $326.37 \mu$Sec and $74.56 \mu$Sec, respectively. The processing time for the novel HetroXCS is $4.4$ times more than the XCSLib. However, this cost is justified because the HetroXCS needs on average $4.3$ steps to successfully reach the goal in Maze7, whereas the XCSLib needs $31$ steps. Thus XCSLib utilizes $7.2$ times more steps as compared to HetroXCS. This shows that the overall computational cost of the XCSLib is more than the HetroXCS. Despite this cost, there is no guarantee that the XCSLib based agent successfully reach the goal, whereas HetroXCS based agent successfully reached the goal in all trials. 

\section{Discussion}
\label{disc}
This work was designed to investigate the benefits/costs of lateralization in cognitive inspired AI systems. The lateralized approach enables the AI systems to consider the given problem instance at different levels of abstraction, i.e. local viewpoint and big picture. Lateralized AI systems have the ability to simultaneously address the elementary features (constituent level) and higher-order abstract features that are extracted across niches (holistic level). This ability empowers the AI systems to efficiently solve complex problems that similar non-lateralized (homogeneous) systems struggle to solve. 

The computer vision problems are used to investigate the robustness of the lateralized approach against adversarial attacks, where the additional benefits are evaluated to include exciting or inhibiting the appropriate BBKs to gain computation efficiency. The adversarial images have redundant, noisy, and irrelevant data. These images proved to be challenging for state-of-the-art (non-lateralized) deep networks. In contrast, the lateralized approach enables the novel lateralized system to correctly classify such images including those that are badly affected by the strong adversarial attacks. The computational cost of a lateralized AI system is double as compared with the conventional AI system. However, this cost is justified based on the robustness exhibited by the lateralized system against adversarial attacks. It is hard to fool the novel lateralized system because it requires an adversarial attack to successfully challenge the constituents as well as the holistic components of an image. The costs of lateralization may dominate in simple computer vision problems where inputs are clean and unambiguous. However, when inputs include noise and irrelevant data (as is common in the real world), the benefits of a lateralised system outweigh the computations costs.

The navigation problems are multi-step path planning problems that provide a virtual environment that simulates real-world navigational problems. These problems are used to investigate the effectiveness of the heterogeneous approach in resolving complex mazes. These mazes entail hierarchical aliasing patterns that pose additional challenges for the homogeneous systems due to their reliance on local viewpoint only. Whereas, the novel approach enables the lateralized AI agent to simultaneously consider the multiple viewpoints of the given environment, the local viewpoint, and the world viewpoint. Consequently, the novel system optimally resolves the complex hierarchical patterns of aliased states that similar homogeneous systems struggle to resolve. The per-step computational cost of the novel system is $4.4$ times higher than the conventional AI system. However, the average number of steps required by the lateralized agent to successfully reach the goal state is $7.2$ times less than the conventional AI agent. Therefore, the overall computational cost of the lateralized system is less than the conventional AI system. Moreover, the lateralized agent successfully reached the goal state in all trials but this is not true for conventional AI agents.

The lateralized AI system may not work better than a conventional AI system for the optimization (or other such) problems where a given problem instance cannot be considered at different levels of abstraction. Moreover, the lateralized approach may not be advantageous if the constituent level BBKs are either unavailable or cannot be (re)utilized to resolve higher (holistic) level problems. The learning pace of a lateralized system may be slow at the start due to extra computations required to learn constituent BBKs, however, once these BBKs are learned, the lateralized system scale quickly to solve complex problems. It is considered that the lateralization is beneficial/advantageous for AI systems in certain situations but there are associated costs as well. This work may open a new door for developing AI systems that can provide insight into contentious topics in neuroscience.

Lateralized AI systems have not been investigated due to the following reasons. (i) A simple problem does not need lateralization. (ii) Lateralized systems need to process the same input signal at the constituent level and holistic level, which appears to double the workload. So a major performance benefit is needed in compensation. (iii) Two different pathways are required to process the same signal at two different levels, i.e. constituent level and holistic level. (iv) It is not always clear how to take a single instance and split it up such that the constituent and holistic level can be considered not just at the same time, but also in a complementary manner. (v) The processing of the same signal at different levels increases the workload. It is necessary to devise a strategy to utilize excite and inhibit signals for the selection of appropriate knowledge structure to avoid extraneous computations. (vi) It is not clear how to store and (re)use learned lateralized knowledge in similar, related, and different domains. (vii) The robustness of lateralized systems, to handle uncertainty and noise in data, is unknown. (viii) It is not clear whether lateralization can be applied to a wide range of problems and scenarios, e.g. single or multiple steps, supervised or reinforcement learning, Boolean or real-valued features, and aliased or non-aliased environments, etc.

\section{Conclusion}
\label{Con}
These two lateralized AI systems highlight the costs and benefits of lateralization in two problem domains. Creating AI systems with lateralization/heterogeneity represents an interesting way to explore the trade-offs and costs of these approaches. The ability to consider the same input signal (problem instance) at different levels of abstraction enables the lateralized AI systems to address constituent (local) features and high-level abstract patterns (features made-up of features) simultaneously. Consequently, the lateralized systems efficiently resolve complex hierarchical patterns by identifying and utilizing the relevant constituent and holistic BBKs. The overall computational costs of the lateralized AI systems are either less than the conventional AI systems or compensated by the associated benefits such as robustness.  

Numeric or real-valued features, supervised or reinforcement learning, single or multiple step, and aliased or non-aliased environments have all been shown to benefit from the lateralized approach in an AI system. The ability to activate/deactivate the most appropriate system module through inhibit or excite signals enables the lateralized systems to reduce costs and offers effectiveness and efficiencies beyond conventional AI approaches. The experimental results demonstrate that lateralized AI systems outperformed similar non-lateralized (homogeneous) AI systems in resolving complex problems in computer vision and navigation domains. It is concluded that the lateralization is beneficial/advantageous for AI systems in certain situations but there are associated costs as well.

\section*{Acknowledgment}
This work is supported by Science for Technological Innovation National Science Challenge, New Zealand.

\bibliographystyle{IEEEtran}

\bibliography{bibAB}
%\bibliography{BibAB4Trans}
%\printbibliography

\ifCLASSOPTIONcaptionsoff
  \newpage
\fi

\vspace{-2.5em}
\begin{IEEEbiography}[{\includegraphics[width=1in,height=1.25in,clip,keepaspectratio]{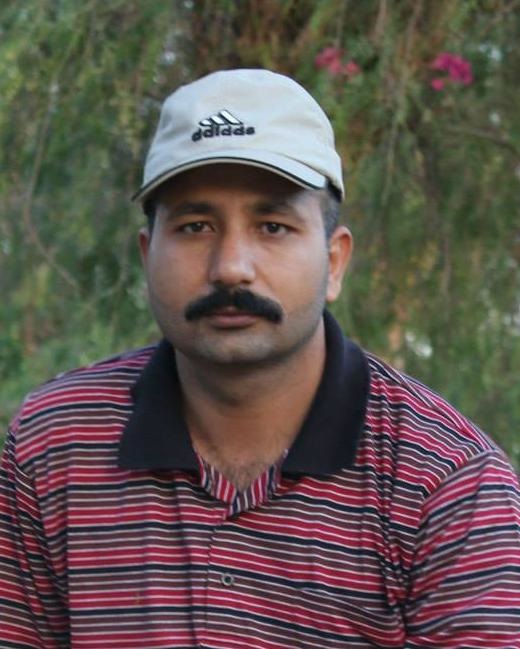}}]{Abubakar Siddique} is a PhD student at School of Engineering and Computer Science, Victoria University of Wellington. His research interests are Learning Classifier Systems, Evolutionary Computing, Brain Lateralization, and Cognitive Neuroscience. Mr Siddique did his undergraduate (major in Computer Science) from Quaid-i-Azam University Islamabad and Master in Computer Engineering from U.E.T Taxila. He was the recipient of ``Student Of The Session'' award. His undergraduate senior project was conducted in an internship at Ultimus, where his work was deployed at the company's Workflow product. He spent nine years at Elixir Pakistan, a California based software Company. His last designation was a Principal Software Engineer where he lead a team of software developers. He developed enterprise level software for customers such as Xerox, IBM and Finis.
%\vspace{-2.0em}
\end{IEEEbiography}
\vspace{-2.5em}
\begin{IEEEbiography}[{\includegraphics[width=1in,height=1.25in,clip,keepaspectratio]{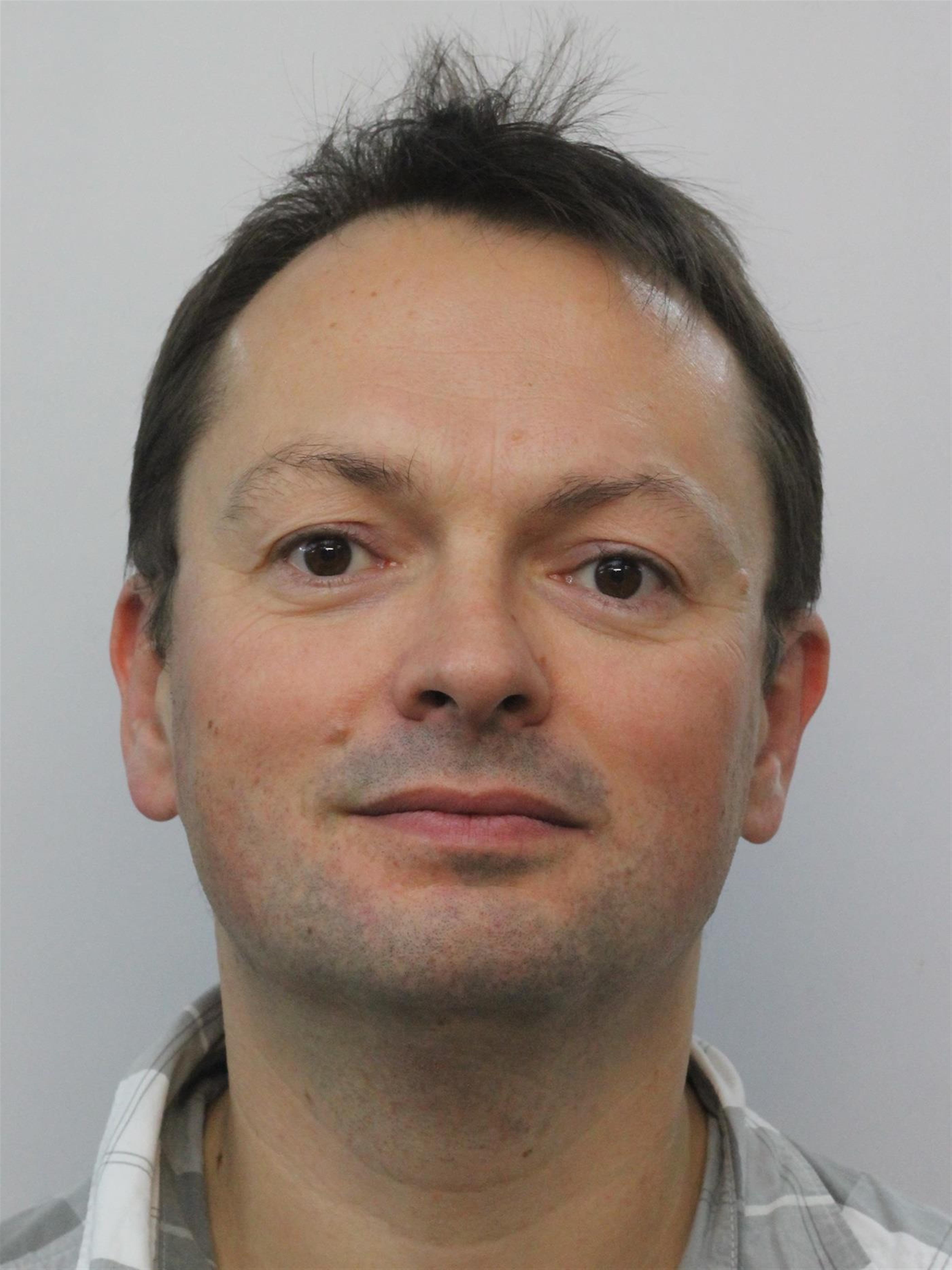}}]{Will N. Browne's} research interests are in developing Artificial Cognitive Systems. His background is in Mechanical Engineering (B.Eng. degree (Hons.)) from the University of Bath, U.K., 1993, and both the M.Sc. degree in Energy and the EngD degree (Engineering Doctorate Scheme) from the University of Wales, Cardiff, U.K., in 1994 and 1999, respectively. After lecturing for eight years at the Department of Cybernetics, University of Reading, Reading, U.K., he is now an Associate Professor with the Evolutionary Computation Research Group, School of Engineering and Computer Science, Victoria University of Wellington, New Zealand. He has published over 100 academic papers in books, refereed international journals, and conferences. This includes two best paper awards at the ACM Genetic and Evolutionary Computation Conference (GECCO), where he has also served as a track chair on four occasions in the Evolutionary Machine Learning Track (or equivalent). He serves on the editorial board of Applied Soft Computing Journal. Together with Dr Ryan Urbanowicz, he has authored the textbook ‘Introduction to Learning Classifier Systems’, Springer, 2017. He was the Co-Local Chair for the IEEE Congress Evolutionary Computation (CEC), Wellington, 2019.
\end{IEEEbiography}
\vspace{-2.5em}
\begin{IEEEbiography}[{\includegraphics[width=1in,height=1.25in,clip,keepaspectratio]{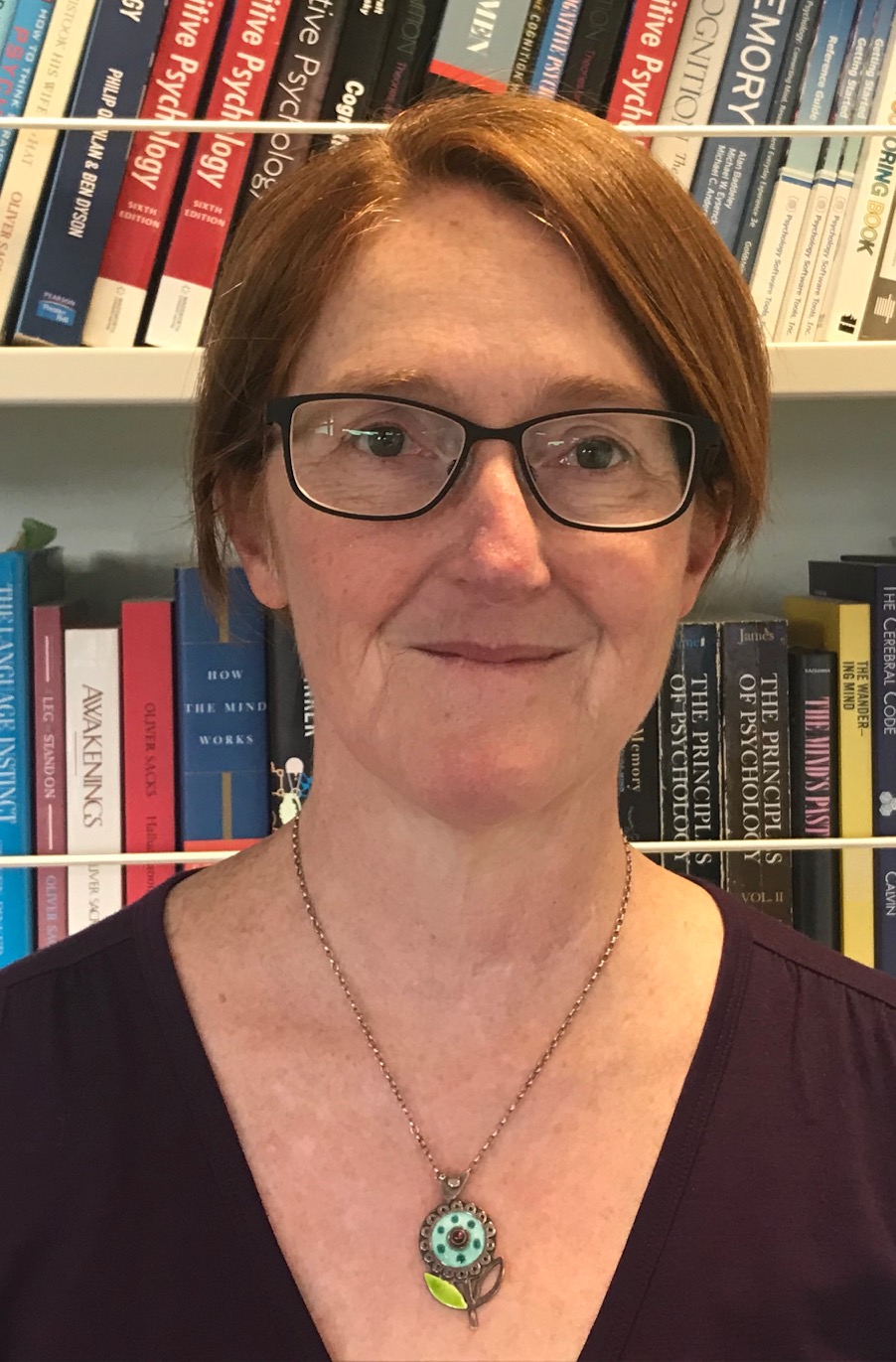}}]{Gina M. Grimshaw} received a BSc in Biochemistry from the University of Toronto in 1987, and a PhD in Cognitive Psychology from the University of Waterloo in 1996. She was a post-doctoral fellow in the Department of Cognitive Science at the University of California San Diego (1996-1997) before taking up an academic position at California State University San Marcos. Since 2007 she has been at Victoria University of Wellington, where she is Associate Professor of Psychology, Programme Director of Cognitive and Behavioural Neuroscience, and Director of the Cognitive and Affective Neuroscience Lab. Her research explores the cognitive and neural mechanisms that support cognition and emotion, with a particular focus on hemispheric specialization and interaction. She has supervised over 20 postgraduate students in Psychology, Cognitive and Behavioural Neuroscience, and Engineering. Her research has been funded by the National Institute of Mental Health (US) and the Royal Society of New Zealand Marsden Fund. She has authored over 50 refereed journal publications, and is Editor of Laterality: Asymmetries of Body, Brain, and Cognition (2016 – present). She is Secretary of the Australasian Society for Experimental Psychology, and chaired the Society’s Experimental Psychology Conference (EPC) in 2011 and 2019. She has won university awards for Teaching Excellence, Research Excellence, and Contributions to Equity and Diversity. 
\end{IEEEbiography}

%\vfill


%\appendices

\end{document}